\documentclass[10pt,journal,compsoc]{IEEEtran}
\usepackage{microtype}
\usepackage{graphicx}
\usepackage{subfigure}
\usepackage{booktabs} % for professional tables
\usepackage{hyperref}
\usepackage{epsfig}
\usepackage{graphicx}
\usepackage{amsmath}
\usepackage{amssymb}
\usepackage{color}
\usepackage[normalem]{ulem}
\usepackage{calligra}
\usepackage{multirow}
\usepackage{verbatim}

\definecolor{mayablue}{rgb}{0.45, 0.76, 0.98}
\definecolor{burntorange}{rgb}{0.8, 0.33, 0.0}

\newcommand{\RR}{\mathbb{R}}
\newcommand{\EE}{\mathbb{E}}
\newcommand{\picwSix}{0.16\linewidth}

\newcommand{\picwFour}{0.245\linewidth}
\newcommand{\picwTwo}{0.45\linewidth}
\newcommand{\mySkip}[1]{{\hskip #1\linewidth}}
\newcommand{\eg}{\textit{e.g.}}
\newcommand{\ie}{\textit{i.e.}}
\newcommand{\etc}{\textit{etc.}}
\newcommand{\etal}{\textit{et al.}}

\ifCLASSOPTIONcompsoc
  \usepackage[nocompress]{cite}
\else
  \usepackage{cite}
\fi
\ifCLASSINFOpdf
\else
\fi
\begin{document}
\title{End-to-end Active Object Tracking\\ and Its Real-world Deployment via \\Reinforcement Learning}

\author{Wenhan Luo*,
        Peng Sun*,
        Fangwei Zhong*,
        Wei Liu,
        Tong Zhang,
        and Yizhou Wang% <-this % stops a space
\IEEEcompsocitemizethanks{
\IEEEcompsocthanksitem{* indicates equal contributions.}
\IEEEcompsocthanksitem Wenhan Luo, Peng Sun, Wei Liu, and Tong Zhang are with Tencent AI Lab, Shenzhen 518057, P.R. China. E-mail: \{whluo.china@gmail.com, pengsun000@gmail.com, wl2223@columbia.edu, tongzhang@tongzhang-ml.org\}
\IEEEcompsocthanksitem Fangwei Zhong and Yizhou Wang are with the Nat'l Eng. Lab. for Video Technology, Key Lab. of Machine Perception (MoE), Computer Science Dept., Peking University, Beijing, 100871, China, Cooperative Medianet Innovation Center, Peng Cheng Lab. E-mail: \{zfw@pku.edu.cn, Yizhou.Wang@pku.edu.cn\} }% <-this % stops an unwanted space
}

\IEEEtitleabstractindextext{%
\begin{abstract}
We study active object tracking, where a tracker takes visual observations (\ie, frame sequences) as input and produces the corresponding camera control signals as output (\eg, move forward, turn left, \etc). 
Conventional methods tackle tracking and camera control tasks separately, and the resulting system is difficult to tune jointly. These methods also require significant human efforts for image labeling and expensive trial-and-error system tuning in the real world. 
To address these issues, we propose, in this paper, an end-to-end solution via deep reinforcement learning. 
A ConvNet-LSTM function approximator is adopted for the direct frame-to-action prediction. 
We further propose an environment augmentation technique and a customized reward function, which are crucial for successful training. 
The tracker trained in simulators (ViZDoom and Unreal Engine) demonstrates good generalization behaviors in the case of unseen object moving paths, unseen object appearances, unseen backgrounds, and distracting objects. 
The system is robust and can restore tracking after occasional lost of the target being tracked.
We also find that the tracking ability, obtained solely from simulators, can potentially transfer to real-world scenarios.
We demonstrate successful examples of such transfer, via experiments over the VOT dataset
and the deployment of a real-world robot using the proposed active tracker trained in simulation.
\end{abstract}

}

\maketitle

\IEEEdisplaynontitleabstractindextext

\IEEEpeerreviewmaketitle

%%%%%%%%%%%%%%%%%%%%%%%%%%%%%%%%%%%%%%%%%%%%%%%%%%%%%%%%%%%%%%%%%%%%%%%%%%%%%%%%%%%%%%
%%%%%%%%%%%%%%%%%%%%%%%%%%%%%%%%%% INTRODUCTION %%%%%%%%%%%%%%%%%%%%%%%%%%%%%%%%%%%%%%
%%%%%%%%%%%%%%%%%%%%%%%%%%%%%%%%%%%%%%%%%%%%%%%%%%%%%%%%%%%%%%%%%%%%%%%%%%%%%%%%%%%%%%
\IEEEraisesectionheading{\section{Introduction}
\label{sec:intro}}

\begin{figure}[t]
\begin{center}
\includegraphics[width=0.95\linewidth]{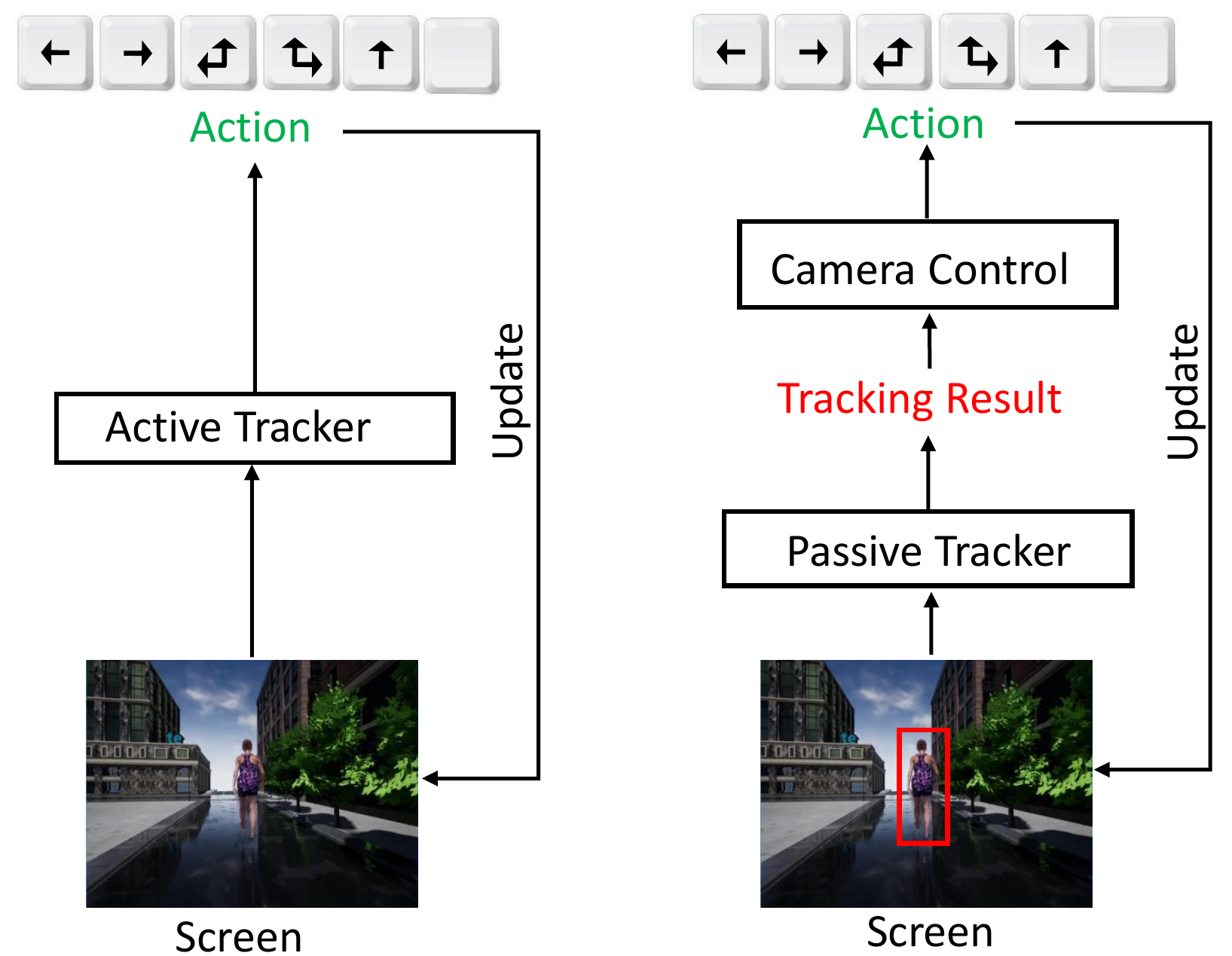}
\end{center}
\caption{The pipeline of active tracking. Left: end-to-end approach. Right: passive tracking plus other modules.}
\label{fig:pipeline}
\end{figure}

\IEEEPARstart{O}{bject} tracking has gained much attention in recent years \cite{bertinetto2016staple,Danelljan_2017_CVPR,zhu2016beyond,cui2016recurrently}. 
The goal of object tracking is to localize an object in continuous video frames given an initial annotation in the first frame. 
Much of the existing work, however, is on \emph{passive} tracker, where it is assumed that the object of interest is always in the image scene, and there is no need to handle camera control during tracking. 
This approach is not suitable for some use-cases, \eg, the tracking performed by a mobile robot with a camera mounted or by a drone. 
For such applications, one should seek a solution to approach \emph{active} tracking, which unifies the two sub-tasks, \ie, the object tracking and the camera control (Fig.~\ref{fig:pipeline}, Right).

In the passive tracker approach, it is difficult to jointly tune the pipeline with the two separate sub-tasks. 
The tracking task may also involve many human efforts for bounding box labeling. 
Moreover, the implementation of camera control is non-trivial and can incur many expensive trial-and-errors system tunings in the real-world. 
To address these issues, we propose an end-to-end active tracking solution via deep reinforcement learning. 
To be specific, we adopt a ConvNet-LSTM network, which takes  raw video frames as input and outputs camera movement actions (\eg, move forward, turn left, \etc). %See Fig. \ref{fig:pipeline}, Left.  

We leverage virtual environments to conveniently simulate active tracking, saving the expensive human labeling or real-world trial-and-error. 
In a virtual environment, an agent (\ie, the tracker) observes a state (a visual frame) from a first-person perspective and takes an action, and then the environment returns the updated state (next visual frame). 
We adopt the modern Reinforcement Learning (RL) algorithm A3C \cite{mnih2016asynchronous} to train the agent, where a customized reward function is designed to encourage the agent to closely follow the object.

We also adopt an environment augmentation technique to boost the tracker's generalization ability. 
For this purpose, much engineering is devoted to preparing various environments in different object appearances, different backgrounds, and different object trajectories. 
We manage this by either using a simulator's plug-in or developing specialized APIs to communicate with a simulator engine. See Sec. \ref{sec:aug-env}.

To our slight surprise, the trained tracker shows good generalization capability. 
In testing, it performs robust active tracking in the case of unseen object movement path, unseen object appearance, unseen background, and distracting object. 
Additionally, the tracker can restore tracking when it occasionally loses the target due to, \eg, abrupt object movement.
See Sec.~\ref{subsec:vizdoom_exp},~\ref{subsec:ue_exp} for details.

In our experiments, the proposed tracking approach also outperforms a few representative conventional passive trackers which are equipped with a hand-tuned camera-control module. 
Although our goal is not to beat the state-of-the-art passive tracker in this work, the experimental results do show that using a passive tracker is not necessary in active tracking. 
Alternatively, a direct end-to-end solution can be effective. 
As far as we know, there has not yet been any attempt to deal with active tracking using an end-to-end approach like this work.

%We perform qualitative evaluation on some video clips taken from the VOT dataset~\cite{VOT_TPAMI}. 
%The results show that the tracking ability, obtained purely from simulators, can potentially transfer to real-world scenarios. 

%The major part of existing works about reinforcement learning are trained and/or tested in virtual environments due to expensive cost or technical difficulty in real environments. The practicability of these works has not been verified as the transfer from virtual to real environments is usually ignored. Besides proving the potential of transferring by testing the proposed method in the VOT dataset mentioned above, we seriously deal with the virtual-to-real problem in this paper. The gap between virtual and real environments is bridged by more advanced environment augmentation techniques, and more appropriate action space. The real robot, a TurtleBot, trained in virtual environment with the proposed environment augmentation techniques and the improved action space, can successfully follow target well. 

We find out that the tracking ability, obtained purely from simulators, can potentially transfer to real-world scenarios.
We demonstrate the transferring in a two-stage experiment. 

First, 
we perform qualitative evaluation on some video clips taken from the VOT dataset~\cite{VOT_TPAMI}. 
Of course, in this case we cannot really control the camera movement as the VOT videos are ``off-line''.
However, we observe that the tracker, trained in simulators and evaluated over VOT videos, is able to output actions that are consistent with the actual camera movements.
See Sec.~\ref{subsec:vot_exp}.

Second, 
we manage to deploy the proposed active tracking in a real-world robot.
We systematically deal with the virtual-to-real problem.
Specifically, the gap between virtual environment and real world is bridged by more advanced environment augmentation techniques and a more appropriately designed action space. 
In this way, 
the real robot, 
a TurtleBot whose tracking model is trained in virtual environment,
can successfully follow a target well in real-world indoor and outdoor scenes.
See Sec.~\ref{subsec:real-world-exp}.

This paper extends our previous conference paper \cite{luo2018end} in several aspects. 
Primarily, the deployment in a real-world robot is delivered.
That is, we train the active tracking model in virtual environments 
and deploy it in a TurtleBot that performs object tracking in the real world. 
Also, several improvements have been proposed to deal with the virtual-to-real gap.
Specifically, 
1) more advanced environment augmentation techniques have been proposed to boost the environment diversity, which improves the transfer ability tailored to the real world. 
2) A more appropriate action space compared with the conference paper is developed, 
and using a continuous action space for active tracking is investigated. 
3) A mapping from the neural network prediction to the robot control signal is established in order to successfully deliver the end-to-end tracking.

%%%%%%%%%%%%%%%%%%%%%%%%%%%%%%%%%%%%%%%%%%%%%%%%%%%%%%%%%%%%%%%%%%%%%%%%%%%%%%%%%%%%%%%%%%%%%
%%%%%%%%%%%%%%%%%%%%%%%%%%%%%%%%%%%%%%%%% RELATED WORK %%%%%%%%%%%%%%%%%%%%%%%%%%%%%%%%%%%%%%
%%%%%%%%%%%%%%%%%%%%%%%%%%%%%%%%%%%%%%%%%%%%%%%%%%%%%%%%%%%%%%%%%%%%%%%%%%%%%%%%%%%%%%%%%%%%%
\section{Related Work}
\label{sec:relatedwork}
As our work is related to object tracking, reinforcement learning, and environment augmentation, 
we briefly review previous work on these topics in the following.

\subsection{Object Tracking}
Object tracking \cite{wu2013online,luo2014multiple} has been conducted
in both passive and active scenarios. As mentioned in Sec. \ref{sec:intro}, passive object tracking has gained more attention due to its relatively simpler problem settings.
In recent years, passive object tracking has achieved a great progress \cite{wu2013online}. 
Many approaches have been proposed to overcome difficulties resulted from issues such as occlusion and illumination variations.
In \cite{ross2008incremental} incremental subspace learning is adopted to update the appearance model of an object and is integrated into a particle filter framework for object tracking. To ensure that the model is updated appropriately, a method for updating the sample mean and a forgetting factor are employed.
Babenko \textit{et al.} \cite{babenko2009visual} employed multiple instance learning to track an object. An algorithm of online multiple instance learning was developed to update the model for adapting to object appearance variations.
Sparse representation was used in \cite{mei2009robust} for visual
tracking. An object candidate was represented as a combination of a
set of templates. Based on a module of careful template updating, the
candidate with the sparsest coefficient is chosen as the target.
In \cite{kalal2012tracking}, Tracking, Learning and Detection (TLD) were integrated into one framework for long-term tracking, where a detection module could re-initialize the tracker once a missing object reappears. Experimental results showed that this method was robust for long-term tracking.
Hu \textit{et al.} \cite{hu2012single} proposed a block-division appearance model for single and multiple object tracking. The blocks were selectively updated in the case of occlusion, avoiding incorrect update of the occluded parts of an object.
Rather than casting visual tracking as a classification problem, Hare \textit{et al.} used a kernelized structured output support vector machine (SVM) to constrain object tracking. By doing so, they avoided converting positions to labels of training samples. 
Correlation filter based object tracking \cite{Valmadre_2017_CVPR,Choi_2017_CVPR} has also achieved a success in real-time object tracking \cite{bolme2010visual,henriques2015high}.
In recent years deep learning has been successfully applied to object tracking \cite{wang2016stct,bertinetto2016fully}. 
For instance, a stacked denoising autoencoder was trained to learn good representations for object tracking in \cite{wang2013learning}. 
Both low-level and high-level representations were adopted to gain both accuracy and robustness \cite{ma2015hierarchical}.       
 
Active object tracking additionally considers camera control compared with traditional object tracking. 
There exists not much research attention in the area of active tracking. 
%Conventional solutions dealt with object tracking and camera control in separate components \cite{denzler1994active,murray1994motion,kim2005detecting}, but these solutions are difficult to tune.
Conventional solutions dealt with object tracking and camera control in separate components \cite{denzler1994active,murray1994motion,kim2005detecting,torkaman2012real,ccelik2017color,das2018stable}. For example, in \cite{ccelik2017color}, object detection is applied to estimate motion and a camera with pan and tilt operations is employed to keep the object in the field of view. However, these solutions are difficult to tune.
Our proposal is completely different from them as it tackles object tracking and camera control simultaneously in an end-to-end manner.

\begin{figure*}[t]
\begin{center}
\includegraphics[width=0.9\linewidth]{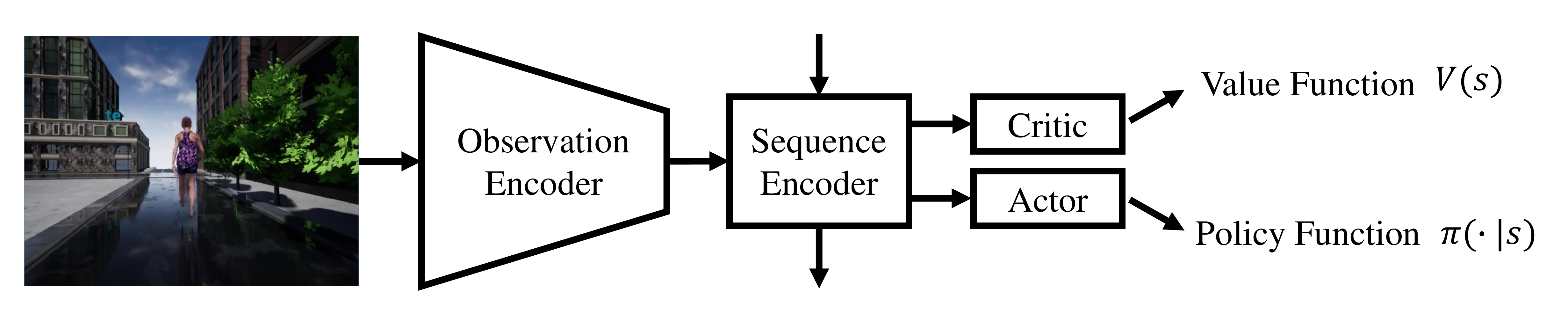}
\end{center}
\caption{An Overview of the network architecture of our active tracker. The observation encoder contains three layers, two convolutional layers and one fully-connected layer. The sequence encoder is a single-layer LSTM, which encodes the image features over time. The actor-critic network corresponds to two branches of fully-connected layers.}
\label{fig:network}
\end{figure*}

%An Overview of the network architecture of our active tracker. The observation encoder contains three layers, two convolutional layers and one fully-connected layer. The sequence encoder is a single-layer LSTM, which encodes the image features over time. The actor-critic network corresponds to two branches of fully-connected layers.

%%%%%%%%%%%%%%%%%%%%%%%  Reinforcement Learning %%%%%%%%%%%%%%%%
\subsection{Reinforcement Learning}
Reinforcement Learning (RL) \cite{sutton1998} is a principled approach to temporal decision making problems.
In a typical RL framework, an \emph{agent} learns from the \emph{environment} a \emph{policy} function that maps a \emph{state} to an \emph{action} at each discrete time step, where the objective is to maximize the accumulated \emph{rewards} returned by the environment. 
Take robot navigation as an example. 
The robot itself is an Agent and its surroundings constitute the Environment. 
The state can be either
high-level (\eg, the position and the speed of the robot) or
low-level (\eg, the raw frame pixels shot from a mounted camera).
The action can be a discrete movement instruction (\eg,
move-forward, turn-left, \etc) and the policy can be an arbitrary
function approximator (\eg, a linear function or a neural
network). The environment returns a negative value as the reward if the
robot hits some obstacle.

Historically, RL has been successfully applied to inventory management, path planning, game playing, \etc.
We refer the readers to \cite{sutton1998} for a thorough survey.

% introduce three kinds of RL framework(Value, policy gradient, actor-critic)

On the other hand, the breakthrough in deep learning in the past few years has advanced computer vision tasks, including image classification \cite{alexnet}, segmentation \cite{long2015}, object detection and localization \cite{rcnn}, and so on. 
In particular, it is believed that deep Convolutional Neural Networks (ConvNets) can learn good feature representations from raw image pixels, which is beneficial to higher-level tasks. 

Equipped with deep ConvNets, RL shows impressive successes on those tasks involving image (-like) raw states, \eg, visuomotor control \cite{levine2016end}, playing board game GO \cite{alphago2016} and video game \cite{atari2015, wu2017}. 
Recently, in the computer vision community there are also preliminary attempts of applying deep RL to traditional tasks, \eg, object localization \cite{caicedo2015} and region proposal \cite{jie2016}. 
There are also methods of visual tracking relying on RL \cite{choi2017visual,Huang_2017_ICCV,Supancic_2017_ICCV,Yun_2017_CVPR}. 
However, they are distinct from our work, as they formulate passive
tracking with RL and do not consider camera controls. In contrast, our focus in this work is active tracking.   

\subsection{Environment Augmentation for Virtual-to-Real} % virtual to real
Using environment augmentation for improving the generalization of a learned model has previously been explored in a large amount of work on robotics, 
including methods for obstacle avoidance~\cite{sadeghi2016rl}, navigation~\cite{wu2018building}, and robot manipulation~\cite{peng2017sim, tobin2017domain, james2017transferring, sadeghi2017sim2real}.
Sadeghi \etal~\cite{sadeghi2016rl} first demonstrated that, agent could learn a collision-free indoor flight policy that generalizes to the real world, by highly randomizing the rendering settings for the simulated training set without a single real image.
A similar technique, 
under the name ``domain randomization'', 
is also widely discussed in the literature of robot manipulation.
In a 3D simulator, the table, plane, and object are rendered with random textures as well as random illuminations, and the physical parameters (\eg, masses and friction parameters) are also randomized.
Domain randomization is shown to produce robust models that can generalize broadly.
The trained model can be directly deployed without fine-tuning in some real world tasks, including obstacle avoidance~\cite{sadeghi2016rl} and simple robot manipulation~\cite{peng2017sim, tobin2017domain, james2017transferring, sadeghi2017sim2real}, even with a model trained in an end-to-end manner~\cite{james2017transferring}.

For visual navigation, Wu \etal~\cite{wu2018building} generated a great number of rooms with different layouts and objects to train a map-free navigator via reinforcement learning.

The closest work to this study is ~\cite{hong2018virtual}, which tries to use a domain randomization method to train an end-to-end tracker but fails in most of real-world scenarios.
To solve this problem, they separate the learning model into a perception module and a control policy module, and use semantic image segmentation as the meta representation to connect these two modules.

Compared to \cite{hong2018virtual}, we augment the environment by randomizing not only the object textures and layouts, but also the motion parameters of the target object, such as velocity and trajectory.
Benefiting from the sequence encoder in our end-to-end model and the environment augmentation method, we successfully deploy our end-to-end active tracker in a real-world robot without fine-tuning in this study.
To the best of our knowledge, this is the first successful demonstration of an end-to-end active tracker trained in virtual environment that can adapt to real world robot settings.
% However, to our knowledge, no prior method has demonstrated that the visual based end-to-end active tracker learned from the virtual environment is able to run in the real-world scenarios.

%However, to the best of our knowledge, no prior method has demonstrated that the visual based end-to-end active tracker learned from the virtual environment is able to run in the real-world scenarios.
%The closed work to this topic has demonstrated that the domain randomization method used in simulation is able to produce robust models that can generalize broadly and can directly be deployed in the real world in robot manipulation tasks~\cite{}. 

%%%%%%%%%%%%%%%%%%%%%%%%%%%%%%%%%%%%%%%%%%%%%%%%%%%%%%%%%%%%%%%%%%%%%%%%%%%%%%%%%%%%%%%%%%%%%%%
%%%%%%%%%%%%%%%%%%%%%%%%%%%%%%%%%%%%%%%%%%%% OUR APPROACH %%%%%%%%%%%%%%%%%%%%%%%%%%%%%%%%%%%%% %%%%%%%%%%%%%%%%%%%%%%%%%%%%%%%%%%%%%%%%%%%%%%%%%%%%%%%%%%%%%%%%%%%%%%%%%%%%%%%%%%%%%%%%%%%%%%%
\section{Our Approach}
\label{sec:approach}
In our approach, virtual tracking scenes are generated for both training and testing. 
To train the tracker, we employ a state-of-the-art reinforcement learning algorithm, A3C \cite{mnih2016asynchronous}. 
For the sake of robust and effective training, we also propose data augmentation techniques and a customized reward function, which are elaborated later.

Although various types of states are available, for the research purpose
in this study we define the state as the RGB screen frame of the first-person view.
To be more specific, the tracker observes the raw visual state and takes one action from the action set $\mathcal{A} = \lbrace$\emph{turn-left, turn-right, turn-left-and-move-forward, turn-right-and-move-forward, move-forward, no-op}$\rbrace$. 
The action is processed by the environment, which returns to the agent the updated screen frame as well as the current reward.

%%%%%%%%%%%%%%% Tracking Scenarios %%%%%%%%%%%%%%%
\subsection{Tracking Scenarios}
\label{sec:tracking_scenario}
It is impossible to train the desired end-to-end active tracker in real-world scenarios. 
Thus, we adopt two types of virtual environments for simulated training. 

\textbf{ViZDoom.} 
ViZDoom \cite{kempka2016vizdoom,ViZDoom} is an RL research platform based on a 3D FPS video game called Doom. 
In ViZDoom, the game engine corresponds to the environment, while the video game player corresponds to the agent. 
The agent receives from the environment a state and a reward at each time step. 
In this study, we make customized ViZDoom maps (see Fig. \ref{fig:map-screen}) composed of an object (a monster) and background (ceiling, floor, and wall). 
The monster walks along a pre-specified path programmed by the ACS
script \cite{kempka2016vizdoom}, and our goal is to train the agent, \ie, the tracker, to follow the object closely.  

\textbf{Unreal Engine.} 
Though convenient for research, ViZDoom does not provide realistic scenarios.
To this end, we adopt Unreal Engine (UE) \cite{unrealengine} to construct nearly real-world environments. 
UE is a popular game engine and has a broad influence in the game industry.
It provides realistic scenarios which can mimic real-world scenes (please see exemplar images in Fig. \ref{fig:unreal-env} and videos in our supplementary materials).
We employ UnrealCV \cite{qiu2017unrealcv}, which provides convenient APIs, along with a wrapper \cite{gymunrealcv2017} compatible with OpenAI Gym \cite{opanai_gym}, for interactions between RL algorithms and the environments constructed based on UE. 

%%%%%%%%%%%%%%%%%%%%%%%%% A3C %%%%%%%%%%%%%%%%%%%%%
\subsection{Reinforcement Learning Algorithms}
\label{sec:actor-critic}
% formulation env observation action 
We consider the standard reinforcement learning setting where an agent interacts with an environment over a number of discrete time steps. 
The observation at time step $t$ is denoted by $o_t$. 
In this study, it corresponds to a raw image taken by a camera in the agent's first-person view.
The state $s_t$ is the experience history up to time $t$, 
%$s_t = h(o_1, r_1, a_1, ..., o_t, r_t)$.
$s_t = (o_1, o_2, ..., o_t)$.
At time $t$, the agent receives an observation $o_t$ from the environment. 
Meanwhile, the agent also receives a reward $r_t \in \RR$ according to a \emph{reward function} $r_t = g(s_t)$, which will be characterized in Sec. \ref{sec:reward-fun}. 
Subsequently, an action $a_t \in \mathcal{A}$ is drawn from the agent's policy function distribution: $a_t \backsim \pi(\cdot|s_t)$,
where $\mathcal{A}$ denotes the set of all possible actions.
The updated state $s_{t+1}$ at next time step $t+1$ is subject to a certain but unknown state transition function $s_{t+1} = f(s_t, a_t)$, governed by the environment.
In this way, we can observe a \emph{trajectory} consisting of a sequence of tuplets $\tau = \lbrace \ldots, \left(s_t, a_t, r_t\right) , \left( s_{t+1}, a_{t+1}, r_{t+1}\right), \ldots \rbrace$. 

Denote by $R_{t:\infty} = r_t + \gamma r_{t+1} + \gamma^2 r_{t+2} + ...$ the discounted accumulated reward. 
The \emph{value function} is the expected accumulated reward in the future given state $s_t$, $V^\pi(s) = \EE [ R_{t:\infty}|s_t = s, \pi ]$.
The \emph{action-value function} $Q^\pi(s, a) = \EE [ R_{t:\infty}|s_t = s, a_t = a, \pi ]$ is the expected return following action $a$ from state $s$. 
The \emph{advantage function} $A^\pi(s,a) = Q^\pi(s, a) - V^\pi(s)$ represents a relative measure of the importance of each action.

%When the agent receives the observation $o_t$, an action $a_t \in \mathcal{A}$ is taken, drawn from a policy function distribution: $a_t \backsim \pi(a|s)$.
%referred to as an \emph{Actor}. 

%The action taken at the same time step $t$ is denoted by $a_t \in \mathcal{A}$

%An action, $a_t \in \mathcal{A}$, is drawn from a policy function distribution: $a_t \backsim \pi(\cdot|s_t) \in \RR^K$, referred to as an \emph{Actor}. 

% value based
\emph{Value-based} reinforcement learning algorithms~\cite{watkins1989learning, sutton1998}, such as deep Q-learning ~\cite{watkins1992q, mnih2015human}, learn the action-value function $Q^\pi(s, a; \theta)$ with a function approximator parametrized by $\theta$, and update the parameters $\theta$ by minimizing a mean-squared error derived from the Bellman iterative equation.

% Policy gradient
\emph{Policy gradient} algorithms~\cite{williams1992simple, sutton2000policy} learn a policy by maximizing the expected accumulated reward, $J_\pi(\theta)=\EE[R_{1:\infty};\pi(a|s;\theta)]$.
By policy gradient theorem, it reduces to updating the parameter $\theta$ with the gradient $\EE[\nabla_{\theta}\log\pi(a|s)h(s, a)]$,
where $h(s, a)$ can be either the discounted accumulated reward $R(s)$ or the advantage function $A(s, a)$ given $h(\cdot)$, leading to an unbiased and variance reduction gradient approximator.

% actor critic
The \emph{Actor-Critic} algorithm~\cite{sutton2000policy, peters2008natural} takes the advantage $A(s,a)$ as the multiplier in the above policy gradient term,
and jointly models the policy function $\pi(a|s)$ and the value function $V^\pi(s)$.
%The policy function $\pi\left( \cdot\right) $ and the value function $V\left( \cdot\right)$ are then jointly modeled by a neural network, as will be discussed in Sec. \ref{sec:network}. 
When using a neural network as the function approximator~\cite{mnih2016asynchronous}, we rewrite them as $\pi(a |s_t;\theta)$ and $V(s_t; \theta')$ with parameters $\theta$ and $\theta'$, respectively. Then, we can learn $\theta$ and $\theta'$ over the trajectory $\tau$ by simultaneous stochastic policy gradient and value function regression:
\begin{small}
\begin{align}
\label{eq:param-update}
	 & \theta \leftarrow  \theta + \alpha \nabla_{\theta} \log\pi(a_t|s_t;\theta) A(s_t, a_t) + \beta \nabla_{\theta} H\big(\pi(a |s_t;\theta)\big),  \\
	 & \theta^{'} \leftarrow  \theta' - \alpha \nabla_{\theta^{'}} \frac{1}{2} \big( R_{t:t+n-1} + \gamma^n V(s_{t+n};\theta^{'-} ) - V(s_t;\theta^{'} ) \big)^2,
\end{align}
\end{small}
\\where $R_{t:t+n-1} = \sum_{i = 0}^{n-1} \gamma^{i} r_{t+i}$ is a discounted sum of $n$-step future rewards with factor $0 < \gamma \le 1$,
$\alpha$ is the learning rate, $H\left( \cdot\right)$ is an entropy regularizer, 
$\beta$ is the regularizer factor, 
$\theta^{'-}$ denotes parameters in previous step, and the optimization is with respect to $\theta^{'}$. 
Here an $n$-step bootstrap is adopted, where the accumulated future reward after step $t+n$ has been truncated and replaced by the value function learnt previously, while the accumulated rewards between $t$ and $t+n$ are collected and calculated from the trajectory.
This way, it permits the parameter updating after every $n$ steps.

% A3C Algorithm
In our approach, we adopt a popular RL algorithm called \emph{Asynchronous advantage actor-critic}(A3C)~\cite{mnih2016asynchronous}.
In A3C many workers are spawned and run in parallel, each holding an agent that interacts with an environment.
However, the neural network parameters are shared across the workers and updated every $n$ time steps asynchronously in a lock-free manner using Eq. \eqref{eq:param-update} from each worker.
For the purpose of exploration, the policy collecting experience during training is stochastic,
\ie, the action is randomly drawn from the policy distribution.
This kind of asynchronous training method is reported to be fast yet stable, leading to an improved generalization \cite{mnih2016asynchronous}.

Later in Sec. \ref{sec:aug-env}, we will introduce environment augmentation techniques to further improve the generalization ability.

\begin{figure}
\begin{center}
\includegraphics[width=1.0\linewidth]{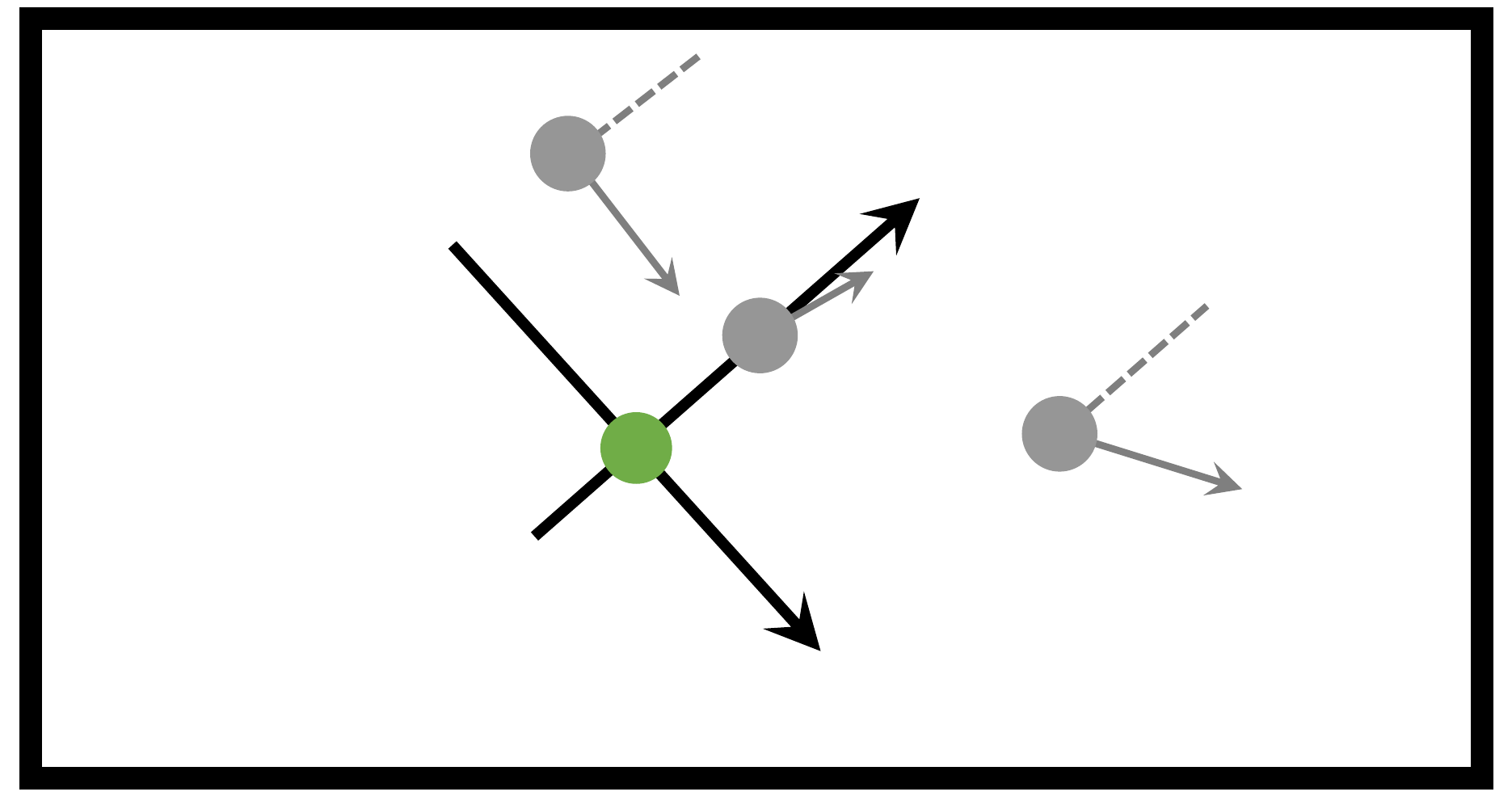}
\end{center}
\caption{A top view of a map with the local coordinate system. The green dot indicates the agent (tracker). The gray dot indicates the initial position and orientation of an object to be tracked. Three gray dots mean three possible initial configurations. Arrow indicates the orientation of an object. Dashed gray lines are parallel to the $y$-axis. The outer thick black rectangle represents the boundary. Best viewed in color.}
\label{fig:coord}
\end{figure}

%%%%%%%%%%%%%%%%%%%%% Network Architecture %%%%%%%%%%%%%%%%%%%
\subsection{Network Architecture}
\label{sec:network}
% TODO:
% 1. formulate as [perception(cnn/cnn+dnn)->sequence relation encoder->actor-critic]
% 2. introduce the choice of each elements(continuous/discrete action for actor, lstm/gru for sequence relation encoder, cnn/ cnn+dnn for perception)
% 3. specify two networks used in experiments
% s_t: raw image    \phi_t: image feature  h_t: recurrent feature \pi: policy fucntion  V:value fucntion 
The active tracker network consists of three primary components: an~\emph{observation encoder}, a \emph{sequence encoder}, and an \emph{actor-critic network}, shown as Fig.~\ref{fig:network}.
The aim of the~\emph{observation encoder} is to extract a feature vector $\phi_t$ for each observed raw image $o_t$. The sequence encoder integrates these image features over time to derive a state representation $h_t$ at each time step $t$. Then the \emph{actor-critic} exploits the hidden state $h_t$ to obtain the policy function $\pi(a; s_t)$ as well as the value function $V(s_t)$.
 
\textbf{Observation Encoder.}
The observation encoder $f_o(s_t)$ transforms the raw pixel observation into a feature vector $\phi_t$ as input to the sequence encoder. Similar to most of deep image encoders, it typically consists of a sequence of convolutional-pooling blocks, extracting information on what is observed in a single image.

\textbf{Sequence Encoder.} % Recurrent Encoder
The sequence encoder $\psi_t = f_s(\phi_1, \phi_2, ..., \phi_t)$ fuses the observation history $\{\phi_1, \phi_2, ... ,\phi_t\}$ to extract a feature representation $\psi_t$, 
which is then fed into the subsequent actor-critic network.

$\psi_t$ models temporal hypotheses about the target motion state over time. 
Note that $\phi_t$ only provides the information about $what$ and $where$ the target is, but not $how$ the target moves.
Consequently, $\phi_t$ alone seems inadequate for our purpose,
because the movement state (velocity, trajectory, \etc) is also intuitively helpful to learn a controlling policy for the active tracking task.

It is intractable to directly feed in the full observation history $\{ \phi_1, \phi_2,...\phi_t \}$.
When building our network architecture, we use a recurrent network as the encoder $f_s(\cdot)$.
In a slight abuse of notations,
we rewrite the sequence encoder as $\psi_t, h_t = f_s(h_{t-1}, \phi_t)$,
where $h_t$ is the hidden state of the recurrent network.
At one single time step $t$, the observation feature vector $\phi_t$ and the hidden state at previous time step $h_{t-1}$ are fed in,
and then the current feature $\psi_t$ and the renewed hidden state $h_t$ are returned. 
Note that $h_{t-1}$ implicitly memorizes the history $\phi_1, ..., \phi_{t-1}$,
capturing ``how'' the object moves.

\textbf{Actor-Critic Network.}
The actor net and the critic net take as input a shared feature $\psi_t$.
The critic approximates the value function $V(s_t)$, which is regarded as the expected future rewards as defined in Sec. \ref{sec:actor-critic}.
The actor outputs the policy distribution $\pi(\cdot; s_t)$ for action decision.
%Referring to Eq.(\ref{eq:param-update}), the value function estimated by the critic is used to approximate the advantage function $A(s,a)$, which is used to compute the policy gradient and optimize the policy function of the actor.
%In general, the critic and the actor are trained jointly with the use of the returned reward $R(a, s_t)$.
The output $V(s_t)$ and $\pi(\cdot; s_t)$ are then used to update the network parameters during training, as in Eq. \eqref{eq:param-update}.

In this study, both the actor and critic network are composed of fully-connected layers.
The output action of the actor can either be continuous or discrete. 
%If the action is in continuous space, each dimension of the action is modeled by Gaussian distribution. 
%So the networks output the mean and the variance of the Gaussian distribution to approximate the policy distribution in continuous space.
If it is continuous, then the action $a_t$ is defined as the agent's velocity in a polar coordinate system $a_t=(\Delta d, \Delta \theta)$,
where $\Delta d$ and $\Delta \theta$ indicate the linear velocity and angular velocity, respectively. 
The policy corresponds to a Gaussian with the mean and standard deviation given by the model output $\pi(\cdot|s_t)$~\cite{mnih2016asynchronous}.
If it is discrete, the policy function will produce a distribution over $K=6$ actions, including ``move-forward", ``turn-left", ``turn-right", ``turn-left-and-move-forward", ``turn-right-and-move-forward", and ``no-op". Note that, in implementation, the compound action, \eg, the action of ``turn-left-and-move-forward" is a combination of actions “turn-left” and “move-forward”, which are provided by the game engine as meta actions.

% If the action is in discrete space, the model output the distribution of the candidate actions.

%At each time step, the actor make a decision to select a action to control the camera to track the target object. The critic aims to approximate a value function $V(s_t)$ which is regarded as the expected long-term rewards. During the training, the value fuction is used to guide the learning of the actor. The output action of the actor can be discrete and continuous.

%%%%%%%%%%%%%%%%%% Reward Function %%%%%%%%%%%%%%%%%%%%%%%
\subsection{Reward Function}
\label{sec:reward-fun}
To perform active tracking, it is a natural intuition that the reward function should encourage the agent to closely follow the object.
In this line of thought, we will first define a two-dimensional local coordinate system, denoted by $\mathcal{S}$ (see Fig. \ref{fig:coord}).
The $x$-axis points from the agent's left shoulder to right shoulder, and the $y$-axis is perpendicular to the $x$-axis and points to the agent's front.
The origin is where the agent is, and the coordinate system $\mathcal{S}$ is parallel to the floor.
Second, we manage to obtain object's local coordinate $(x, y)$ and orientation $\omega$ (in radius) with regard to system $\mathcal{S}$.

With a slight abuse of notations, we can now write the reward function as 
\begin{equation}
\label{eq:reward}
	r = A - \left( \frac{\sqrt{x^2 + (y - d)^2}}{c} + \lambda |\omega| \right),
\end{equation}
where $A > 0$, $c > 0$, $d > 0$, $\lambda > 0$ are tuning parameters. In plain English, Eq. (\ref{eq:reward}) says that the maximum reward $A$ is achieved when the object stands perfectly in front of the agent with a distance $d$ and exhibits no rotation (see Fig. \ref{fig:coord}). $c$ plays a role like a normalization factor to normalize the distance.

In Eq. \eqref{eq:reward} we have omitted the time step subscript $t$ without loss of clarity. 
Also note that the reward function defined this way does not explicitly depend on the raw visual state. Instead, it depends on certain internal states. 
Thanks to the APIs provided by virtual environments, we are able to access the interested internal states and develop the desired reward function.

%\begin{figure}[t]
%%\rule{\textwidth}{1pt}\\
%\hspace*{0.01\linewidth} RandomizedSmall   \mySkip{0.075} CacoDemon  \mySkip{0.065} SharpTurn \\
%\includegraphics[width=\picwThree]{figures/map-small.pdf}
%%\includegraphics[width=\picwSix]{figures/map-standard.pdf}
%%\includegraphics[width=\picwSix]{figures/map-standard.pdf}
%\includegraphics[width=\picwThree]{figures/map-standard.pdf} 
%\includegraphics[width=\picwThree]{figures/map-hard.pdf}\\
%%\includegraphics[width=\picwSix]{figures/map-noise2.pdf} \\
%\includegraphics[width=\picwThree]{figures/screen-small.png}
%%\includegraphics[width=\picwSix]{figures/screen-standard.png}
%\includegraphics[width=\picwThree]{figures/screen-cacodemon.png}
%%\includegraphics[width=\picwSix]{figures/screen-floorceil.png}
%\includegraphics[width=\picwThree]{figures/screen-hard.png}\\
%%\includegraphics[width=\picwSix]{figures/screen-noise2.png}\\
%\vspace{-0.4cm}
%\caption{Maps and screenshots of ViZDoom environments. In all maps, the green dot (with white arrow indicating orientation) represents the agent. The gray dot indicates the object. Blue lines are planned paths and black lines are walls. Best viewed in color.}
%\label{fig:map-screen}
%\vspace{-0.4cm}
%\end{figure}

\begin{figure*}[t]
%\rule{\textwidth}{1pt}\\
\hspace*{0.01\linewidth} RandomizedSmall \mySkip{0.03} Standard  \mySkip{0.075} CacoDemon \mySkip{0.05} FloorCeiling \mySkip{0.065} SharpTurn \mySkip{0.08} Noise2\\
\includegraphics[width=\picwSix]{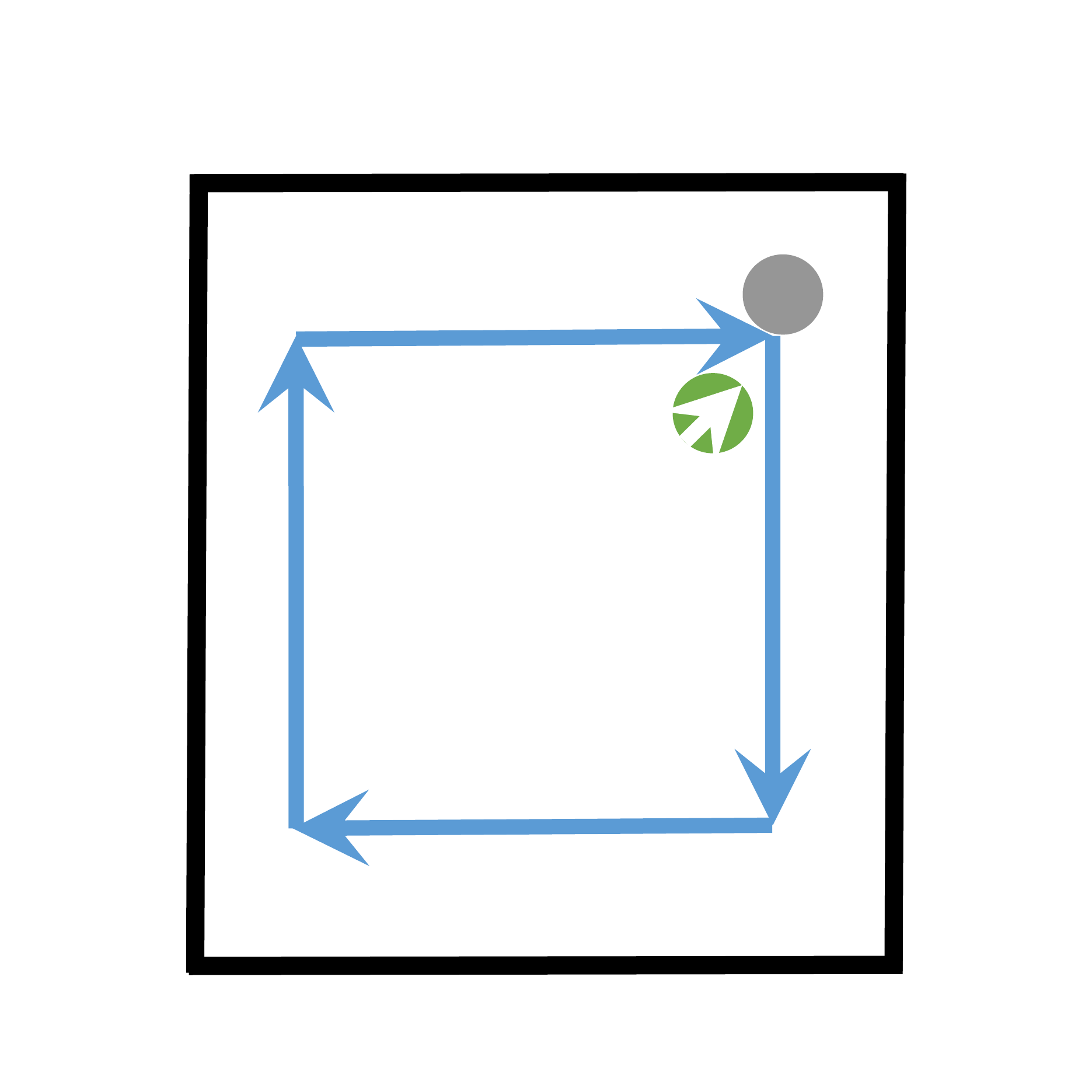}
\includegraphics[width=\picwSix]{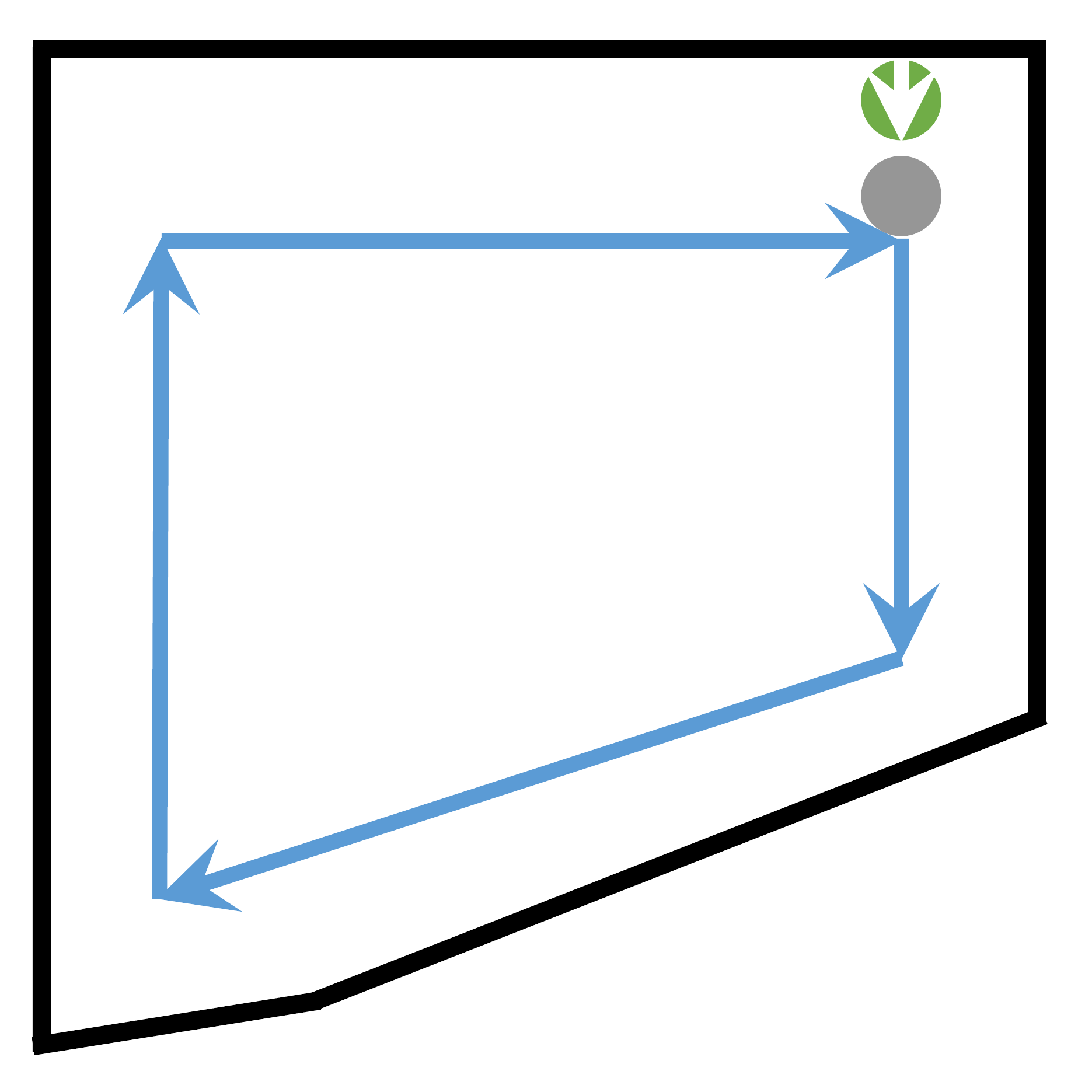}
\includegraphics[width=\picwSix]{figures/map-standard.pdf}
\includegraphics[width=\picwSix]{figures/map-standard.pdf} 
\includegraphics[width=\picwSix]{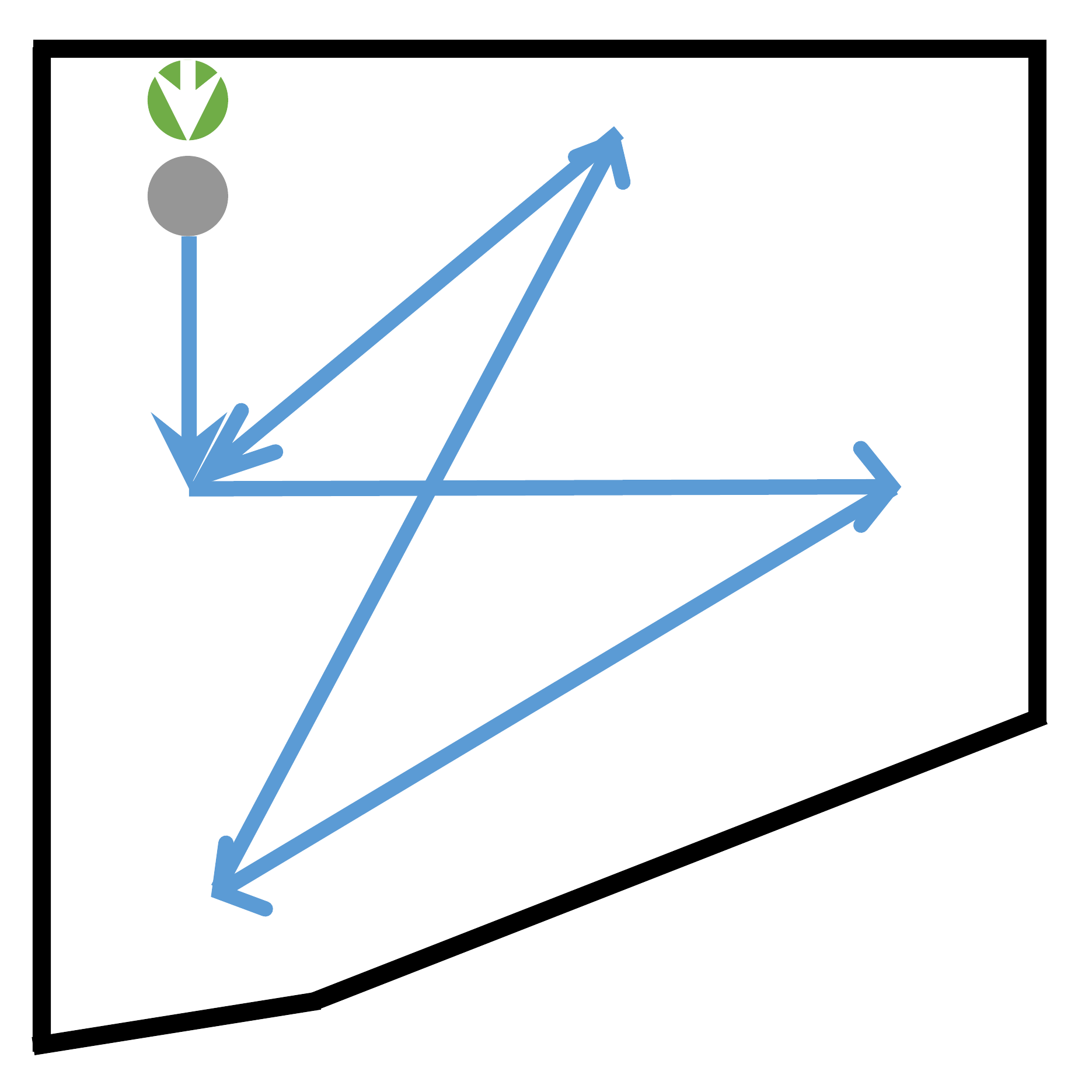}
\includegraphics[width=\picwSix]{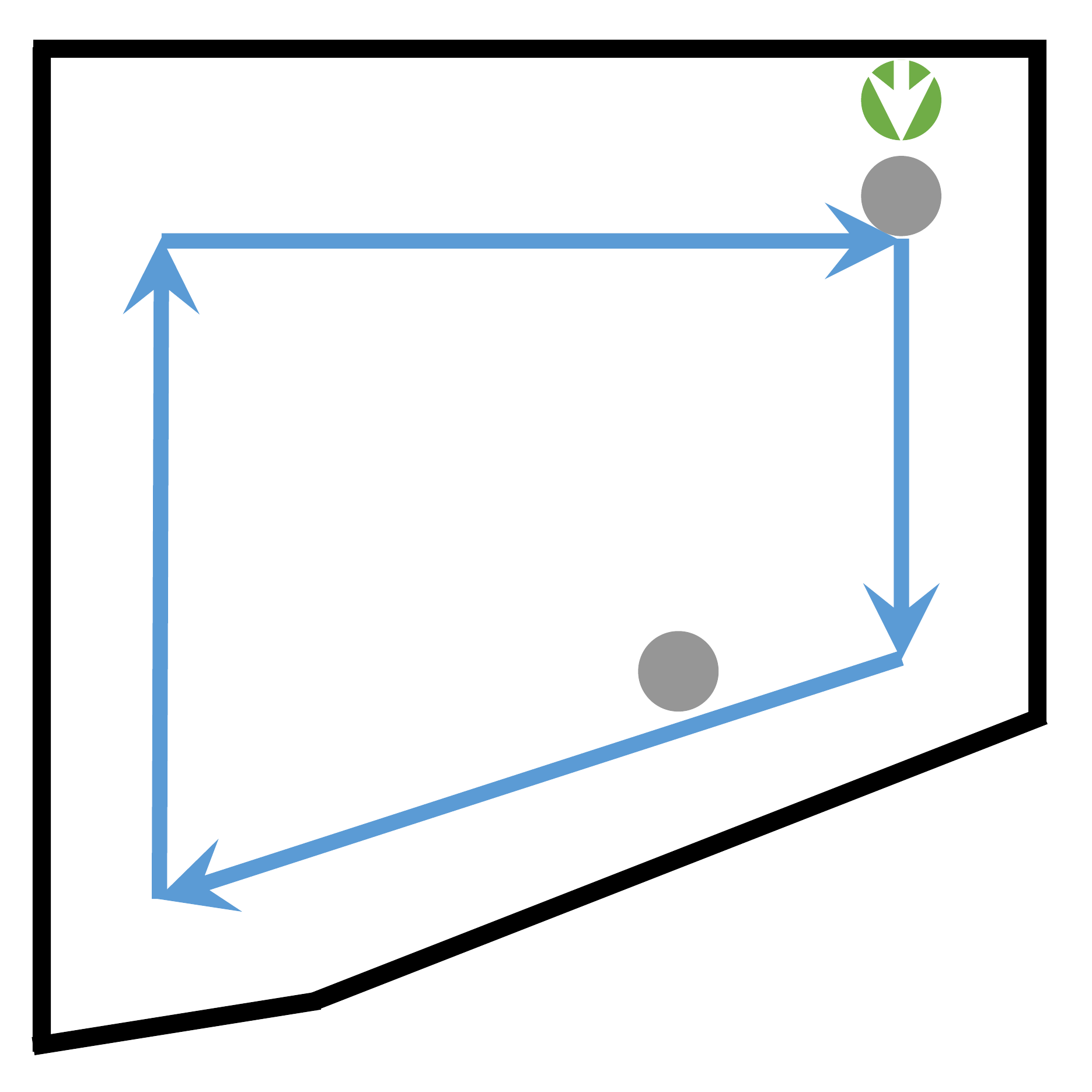} \\
\includegraphics[width=\picwSix]{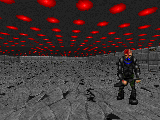}
\includegraphics[width=\picwSix]{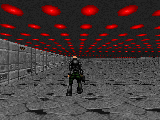}
\includegraphics[width=\picwSix]{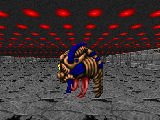}
\includegraphics[width=\picwSix]{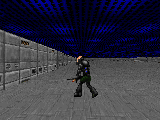}
\includegraphics[width=\picwSix]{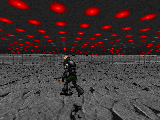}
\includegraphics[width=\picwSix]{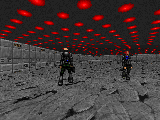}\\
\caption{Maps and screenshots of ViZDoom environments. In all maps, the green dot (with white arrow indicating orientation) represents the agent. The gray dot indicates the object. Blue lines are planned paths and black lines are walls. Best viewed in color.}
\label{fig:map-screen}
\end{figure*}

\begin{figure*}[t]
\begin{center}
\includegraphics[width=1.0\linewidth]{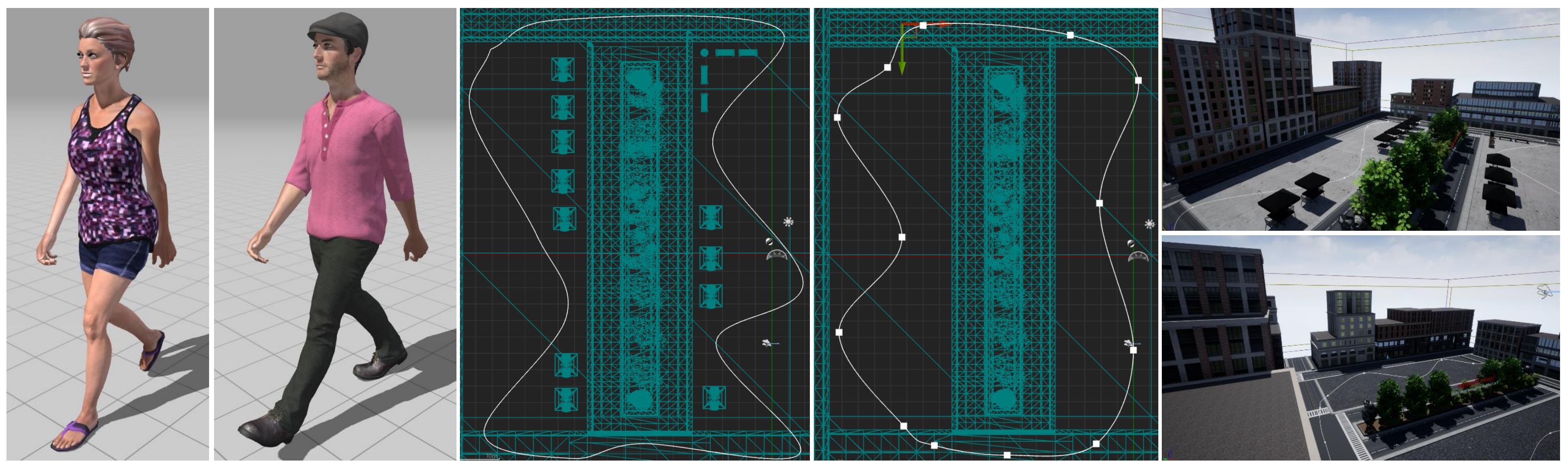}\\
\end{center}
\caption{Screenshots of UE environments. From left to right, there are \emph{Stefani}, \emph{Malcom}, \emph{Path1}, \emph{Path2}, \emph{Square1}, and \emph{Square2}. Best viewed in color.}
\label{fig:unreal-env}
\end{figure*}

%%%%%%%%%%%%%%%%%%%%%%%%  Environment Augmentation %%%%%%%%%%%%%%%%%%%%%%
\subsection{Environment Augmentation}
\label{sec:aug-env}
To make the tracker generalize well, we propose simple yet effective techniques for environment augmentation during training. 

For ViZDoom, recall the object's local position and orientation $(x, y, \omega)$ in system $\mathcal{S}$ described in Sec. \ref{sec:reward-fun}.
For a given environment (\ie, a ViZDoom map) with initial $(x, y, \omega)$,  we randomly perturb it $N$ times by editing the map with the ACS script \cite{kempka2016vizdoom}, yielding a set of environments with varied initial positions and orientations $\{x_i, y_i, \omega_i\}_{i=1}^N$.
We further allow flipping left-right the screen frame (and accordingly the left-right action).
As a result, we obtain $2N$ environments out of one environment.
See Fig. \ref{fig:coord} for an illustration of several possible initial positions and orientations in the local system $\mathcal{S}$.
During the A3C training, we sample uniformly at random one of the $2N$ environments at the beginning of every episode.
As will be seen in Sec. \ref{subsec:vizdoom_exp}, this technique significantly improves the generalization ability of the tracker.

For UE, we construct an environment with a character/target walking along a fixed path.
To augment the environment, we randomly choose some background objects (\eg, tree or building) in the environment and make them invisible.
At the same time, every episode starts from the position, where the agent fails at the last episode. 
This makes the environment and the starting point different from episode to episode, so the variations of the environment during training are augmented.

%%%%%%%%%%%%%%%%%%%%%%%%%%%%%%%%%%%%%%%%%%%%%%%%%%%%%%%%%%%%%%%%%%%%%%%%%%%%%%%%%%%%%%
%%%%%%%%%%%%%%%%%%%%%%%EXPERIMENTAL RESULTS %%%%%%%%%%%%%%%%%%%%%%%%%%%%%%%%%%%%%%%%%%
%%%%%%%%%%%%%%%%%%%%%%%%%%%%%%%%%%%%%%%%%%%%%%%%%%%%%%%%%%%%%%%%%%%%%%%%%%%%%%%%%%%%%%
\section{Experimental Results}
\label{sec:experiment}
The settings are described in Sec. \ref{sec:settings}. 
The experimental results are reported for the virtual environments ViZDoom (Sec. \ref{subsec:vizdoom_exp}) and UE (Sec. \ref{subsec:ue_exp}). 
Qualitative evaluation is performed for real-world sequences from the VOT dataset (Sec. \ref{subsec:vot_exp}) to investigate the transfer potential from virtual environment to real world scenario. In Sec. \ref{subsec:real-world-exp}, we present how we deploy the active tracking algorithm in a real-world robot, showing its practical values.

%%%%%%%%%%%%%%%%%% Settings %%%%%%%%%%%%%%%%
\subsection{Settings}
\label{sec:settings}

\textbf{Environment.}
A set of environments are produced for both training and testing.
For ViZDoom, we adopt a training map as shown in Fig. \ref{fig:map-screen}, left column. 
This map is then augmented as described in Sec. \ref{sec:aug-env} with $N=21$, leading to $42$ environments that we can sample from during training.
For testing, we make other 9 maps, some of which are shown in Fig. ~\ref{fig:map-screen}, middle and right columns. In all maps, the path of the target is pre-specified, indicated by the blue lines. 
However, it is worth noting that the object does not strictly follow the planned path.
Instead, it sometimes randomly moves in a ``zig-zag'' way during the course, which is a built-in game engine behavior.
This poses an additional difficulty to the tracking problem.

For UE, we generate an environment named \emph{Square} with random invisible background objects and a target named \emph{Stefani} walking along a fixed path for training. 
For testing, we make another four environments named as \emph{Square1StefaniPath1 (S1SP1)},~\emph{Square1MalcomPath1 (S1MP1)},~\emph{Square1StefaniPath2 (S1SP2)}, and~\emph{Square2MalcomPath2 (S2MP2)}.
As shown in Fig. \ref{fig:unreal-env}, \emph{Square1} and \emph{Square2} are two different maps, \emph{Stefani} and \emph{Malcom} are two characters/targets, and \emph{Path1} and \emph{Path2} are different paths.
Note that, the training environment \emph{Square} is generated by hiding some background objects in \emph{Square1}. The scenery is changed comparing \emph{S1SP1} with the training environment. From \emph{S1SP1} to \emph{S1MP1}, the target changes. From \emph{S1SP1} to \emph{S1SP2}, the path is different. By adopting the \emph{S2MP2}, we test the performance of tracking when the scenery, target, and path are all varied. We believe that this kind of setting is sufficient to demonstrate the comparison.

For both ViZDoom and UE, we terminate an episode when either the accumulated reward drops below a threshold or the episode length reaches a maximum number.
In our experiments, we let the reward threshold be -450 and the maximum length be 3000, respectively.

\textbf{Metric.}
Two metrics are employed for the experiments.
Specifically, \emph{Accumulated Reward} (AR) and \emph{Episode Length} (EL) of each episode are calculated for quantitative evaluation. 
Note that, the immediate reward defined in Eq. (\ref{eq:reward}) measures the goodness of tracking at some time step, so the metric AR is conceptually much like \emph{Precision} in the conventional tracking literature. 
Also note that too small AR means a failure of tracking and leads to a termination of the current episode. 
As such, the metric EL roughly measures the duration of good tracking, which shares the same spirit as the metric \emph{Successfully Tracked Frames} in conventional tracking applications. 
When letting $A = 1.0$ in Eq. (\ref{eq:reward}), we have that the theoretically maximum AR and EL are both 3000 due to our episode termination criterion. 
In all the following experiments, 100 episodes are run to report the mean and standard deviation, unless specified otherwise.

\begin{table}
\caption{Performance under different protocols in the \emph{Standard} testing environment.}
\label{table:diff_protocol}
\centering
{\begin{tabular}{c c c c}
\hline
\textbf{Protocol}       & \textbf{AR} & \textbf{EL}\\
\hline
\emph{RandomizedEnv}    &2547$\pm$58  &2959$\pm$32  \\
\emph{SingleEnv}        &840$\pm$602  &2404$\pm$287 \\
\hline
\emph{RandomizedEnv (deeper)}    &2535$\pm$64  &3000$\pm$0  \\
\emph{RandomizedEnv (gru)}    &2022$\pm$71   & 3000$\pm$0 \\
\emph{RandomizedEnv (w/o LSTM)}    &1522$\pm$946   & 2693$\pm$702 \\
\hline
\end{tabular}
}
\end{table}

\textbf{Network.}
%The tracker is a ConvNet-LSTM neural network as shown in Fig. \ref{fig:network},
The neural network architecture for the tracker is specified in the following table:

\begin{scriptsize}
\raggedright
\begin{tabular}{c|c|c|c|c|c}
\hline
Layer\# & 1 & 2 & 3 & 4 & 5 \\
\hline
\multirow{2}{*}{Parameters} & \multirow{2}{*}{C8$\times$8-16\emph{S}4} & \multirow{2}{*}{C4$\times$4-32\emph{S}2} & \multirow{2}{*}{FC256} & \multirow{2}{*}{LSTM256} & FC6  \\
\cline{6-6}
 & & & & & FC1 \\
 \hline
\end{tabular}
\end{scriptsize}
where C8$\times$8-16S4 means 16 filters of size 8$\times$8 and stride 4, FC256 indicates dimension 256, and LSTM256 indicates that all the sizes in the LSTM unit are 256.

The observation encoder contains three layers, two convolutional layers, and one fully-connected layer.
The observation $o_t$ is resized to an $84 \times 84 \times 3$  RGB image and fed into the network as input.
The convolutional layers extract features from raw image pixels and the fully-connected layer further transforms the feature representations into a 256-dimensional feature vector $\phi_t$.
Each layer is activated by a ReLU function.

The sequence encoder is a single-layer LSTM with 256 units, which encodes the image features over time. Its output at the last time step $h_{t-1}$ could be used as part of the input, so that the information can propagate as the network recurs over step by step.

The actor-critic network corresponds to two branches of fully-connected layers, FC6, and FC1.
The FC1 indicates a single fully-connected layer with one-dimensional output, which corresponds to the value function $V(s_t)$.
The FC6 indicates a single fully-connected layer with six-dimensional output, which corresponds to the 6-dim discrete policy $\pi\left( \cdot|s_t\right)$.

\textbf{Implementation Details.}
\label{subsec:implementation}
The network is trained from scratch, without pretraining.
The network parameters are updated with Adam optimization, with the initial learning rate $\alpha = 0.0001$. 
The regularizer factor $\beta = 0.01$ and the reward discount factor $\gamma = 0.99$. 
The parameter updating frequency $n$ is 20, and the maximum global iteration for training is $100\times10^6$. 
Validation is performed every 70 seconds and the best validation network model is applied to report performance in testing environments.

%%%%%%%%%%%%%%%%%%% Active Tracking in the ViZDoom Environment %%%%%%%%%%%%%%%
\subsection{Active Tracking in The ViZDoom Environment}
\label{subsec:vizdoom_exp}
We first test the active tracker in a testing environment named \emph{Standard}, showing the effectiveness of the proposed environment augmentation technique. 
The second part of our experiments has more challenging testing environments which vary from the \emph{Standard} environment with regard to object appearance, background, path, and object distraction. 
Comparison with a set of traditional trackers is conducted in the third part, and ablation study of what the tracker has learned is carried out in the last part.

\textbf{Standard Testing Environment.}
In Table \ref{table:diff_protocol}, we report the results in an independent testing environment named \emph{Standard} (see supplementary materials for its detailed description), where we compare two training protocols: with (called~\emph{RandomizedEnv}) or without (called~\emph{SingleEnv}) the augmentation technique as in Sec.~\ref{sec:aug-env}. 
As can be seen,~\emph{RandomizedEnv} performs significantly better than \emph{SingleEnv}. 

\begin{figure*}[t]
\begin{center}
\includegraphics[width=1.0\linewidth]{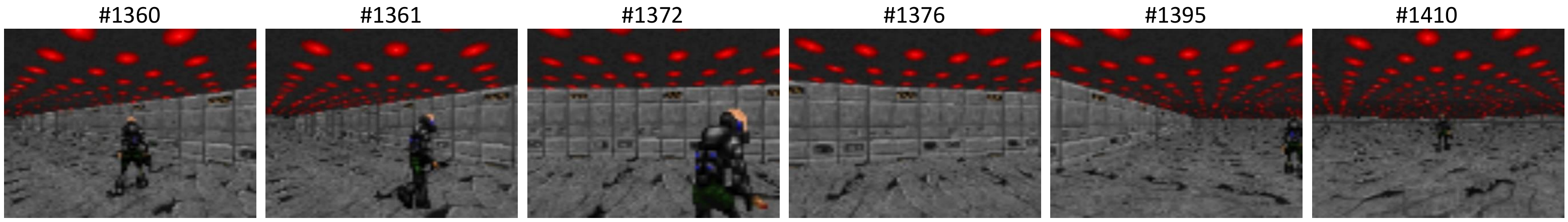}
\end{center}
\caption{Recovering tracking when the target disappears in the \emph{SharpTurn} environment.}
\label{fig:sharpturn}
\end{figure*}

\begin{table}[t]
\caption{Performance of the proposed active tracker in different testing environments.}
\label{table:diff_cases}
\centering
{\begin{tabular}{c c  c}
\hline
\textbf{Environment} & \textbf{AR} & \textbf{EL}\\
\hline
\emph{CacoDemon} & 2415$\pm$71  & 2981$\pm$10 \\
\emph{Zombie}    & 2386$\pm$86  & 2904$\pm$40 \\
\hline
\emph{FloorCeiling} & 1504 $\pm$ 158  & 2581 $\pm$ 84 \\
\emph{Corridor}     & 2636 $\pm$ 34   & 2983 $\pm$ 17 \\
\hline
\emph{SharpTurn}        & 2560$\pm$34  &2987$\pm$12 \\
\emph{Counterclockwise} & 2537$\pm$58  &2964$\pm$23 \\
\hline
\emph{Noise1} & 2493$\pm$72   &2977$\pm$14 \\
\emph{Noise2} & 2313$\pm$103  &2855$\pm$56 \\
\hline
\end{tabular}
}
\end{table}

We discover that the \emph{SingleEnv} protocol quickly exhausts the training capability and obtains the best validation result at about $9 \times 10^6$ training iterations.
On the contrary, the best validation result of \emph{RandomizedEnv} protocol occurs at $48 \times 10^6$, showing that the capacity of the network is exploited better despite the longer training time.
In the following experiments, we only report experimental results with the \emph{RandomizedEnv} protocol.

To verify the importance of the LSTM module, we conduct an ablation study by removing the LSTM, with other settings fixed. The results shown in Table \ref{table:diff_protocol} degrade with a considerable drop (see the AR value), suggesting the LSTM module contributes significantly to the success of the active tracker. We also carry out another study by replacing LSTM with GRU. Results in Table \ref{table:diff_protocol} indicate different RNN modules achieve comparable performance.

We also try a deeper architecture which has three additional convolutional layers. Results are shown in Table \ref{table:diff_protocol}. Comparing the deeper one with the shallow one, the improvement is limited considering longer inference time at the cost of more computations. Therefore, we adopt the shallow architecture throughout the experiment study.

\textbf{Various Testing Environments.} 
To evaluate the generalization ability of our active tracker, we test it in eight more challenging environments as in Table ~\ref{table:diff_cases}. 
These environments are obtained by modifying the \emph{Standard} environment in the following aspects:

1) Change the appearance of the target. Specifically, we have two environments named \emph{CacoDemon} and \emph{Zombie} with targets \emph{CacoDemon} and \emph{Zombie}, respectively. 

2) Revise the background in the environment. We have an environment named \emph{FloorCeiling} with different textures in ceiling and floor, and an environment named \emph{Corridor} with a corridor structure. 

3) Modify the path. The path in \emph{SharpTurn} is composed of several sharp acute angles while the clockwise path is changed to a counterclockwise one in \emph{Counterclockwise}. 

4) Add distractions. \emph{Noise1} is formed by placing a same monster (stationary) near the path along which the target walks. \emph{Noise2} is almost the same as \emph{Noise1}, except that the distracting monster is closer to the path.

From the four categories in Table~\ref{table:diff_cases} we have findings below.\\

%1) The tracker generalizes well in the case of target appearance changing (\emph{Zombie}, \emph{Cacodemon}).\\

%2) The tracker is insensitive to background variations such as changing the ceiling and floor (\emph{FloorCeiling}) or placing additional walls in the map (\emph{Corridor}).\\

%3) The tracker does not lose a target even when the target takes several sharp turns (\emph{SharpTurn}). Note that in conventional tracking, the target is commonly assumed to move smoothly.\\ 

1) The first set of environments aims to test the sensitivity of our tracker to target appearance variations. Even we replace the target monster with completely different targets (\emph{Zombie} and \emph{Cacodemon}), the corresponding results show that it generalizes well in the case of appearance changes.\\

2) The purpose of the second set is to investigate how the tracker works when the background changes. When we change the ceiling, the floor (\emph{FloorCeiling}), or even place additional walls in the map (\emph{Corridor}), the results are not sensitive to background variations.\\

3) In traditional tracking, objects are commonly assumed to move smoothly. The case of abrupt motions is seldom considered. To this end, we also examine the tracker in the \emph{SharpTurn} environment. Even in the case of very sharp turns which are abnormal in practical scenes, the tracker can still chase the object tightly.

We also observe that the tracker can recover tracking when it accidentally loses the target. 
As shown in Fig. \ref{fig:sharpturn}, the target turns right suddenly from frame \#$1360$ to frame \#$1361$. 
Consequently, the tracker loses the target (see frames \#$1372$ and \#$1376$). 
From frame \#$1376$ to \#$1394$, although the target completely disappears in the image, the tracker takes a series of turn-right actions, until frame \#$1395$, where the tracker discovers the target and then continues to track steadily afterwards (frame \#$1410$).
We believe that this capability attributes to the LSTM unit which takes into account historical states when producing current outputs.

%Our tracker performs well when the target walks counterclockwise (\emph{Counterclockwise}), indicating that the tracker does not work by simply memorizing the turning pattern.

In the \emph{Counterclockwise} environment, the tracker tracks well when the object walks along a counterclockwise path, which is not present in the training environment. 
This indicates that the trained tracker does not track the object by memorizing the turning direction.

%4) The tracker is insensitive to a distracting object (\emph{Noise1}), even when the ``bait'' is very close to the path (\emph{Noise2}). 
4) The last set intends to confuse the tracker with bait.
In \emph{Noise1}, the tracker ignores the bait and stably focuses on the object of interest. 
In \emph{Noise2}, though the object acting as a bait is placed very close to the position the target object
will pass, our tracker does not drift to the distraction and consistently focuses on the target object.

%The proposed tracker shows satisfactory generalization in various unseen environments. Readers are encouraged to watch more result videos provided in our supplementary materials.
The results in various testing environments reveal that the proposed tracker does not overfit to specific appearance, background, or path. 
It is even robust to distraction. 
We believe that this benefits from the representation learned from the active ConvNet-LSTM network trained via reinforcement learning. 
Readers are encouraged to watch more result videos provided in our supplementary materials.

\begin{table}[t]
\caption{Comparison with traditional trackers. The best results are shown in bold.}
\label{table:comparison_thirdparty}
\centering
{\begin{tabular}{c c c c}
\hline
\textbf{Environment} & \textbf{Tracker} & \textbf{AR} & \textbf{EL}\\
\hline
\multirow{5}{*}{\emph{Standard}}
& MIL         & -454.2 $\pm$ 0.3  & 743.1 $\pm$ 21.4 \\
& Meanshift   & -452.5 $\pm$ 0.2  & 553.4 $\pm$ 2.2  \\
& KCF         & -454.1 $\pm$ 0.2  & 228.4 $\pm$ 5.5  \\
& Correlation & -453.6 $\pm$ 0.2 & 252.7 $\pm$ 16.6 \\
& TLD & -460.3 $\pm$ 0.3 & 359.7 $\pm$ 10.6 \\
\cline{2-4}
& Active      & \textbf{2457$\pm$58}  & \textbf{2959$\pm$32}  \\
\hline
\multirow{5}{*}{\emph{SharpTurn}}& MIL & -453.3 $\pm$ 0.2   & 388.3 $\pm$ 15.5 \\
& Meanshift & -454.4 $\pm$ 0.3 & 250.1 $\pm$ 1.9\\
& KCF &  -452.4 $\pm$ 0.2  & 199.2 $\pm$ 5.7 \\
&Correlation & -453.0 $\pm$ 0.2 & 186.3 $\pm$ 6.0\\
& TLD & -454.9 $\pm$ 0.4 & 261.9 $\pm$ 2.6 \\
\cline{2-4}
&Active      & \textbf{2560$\pm$34}     & \textbf{2987$\pm$12}  \\
\hline
\multirow{5}{*}{\emph{Cacodemon}}& MIL & -453.5 $\pm$ 0.2   & 540.6 $\pm$ 18.2 \\
& Meanshift & -452.9 $\pm$ 0.2  & 484.3 $\pm$ 9.4\\
& KCF &  -454.5 $\pm$ 0.3  &  263.1 $\pm$ 6.2 \\
&Correlation & -453.3 $\pm$ 0.2 & 155.8 $\pm$ 1.9\\
&TLD & -453.2 $\pm$ 0.4 & 450.6 $\pm$ 17.8 \\
\cline{2-4}
&Active & \textbf{2451$\pm$71}    & \textbf{2981$\pm$10}  \\
\hline
\end{tabular}
}
\end{table}

\textbf{Comparison with Simulated Active Trackers.} 
In a more extensive experiment we compare the proposed tracker with a few traditional trackers.
These trackers are originally developed for passive tracking applications. Particularly, the MIL \cite{babenko2009visual}, Meanshift \cite{comaniciu2000real}, KCF \cite{henriques2015high}, Correlation \cite{danelljan2014accurate}, and TLD \cite{kalal2012tracking} trackers are employed for comparison. These passive trackers are selected as they are typical ones in the development of visual tracking. Some of them are indeed outdated. However, we did not intend to defeat the state-of-the-art passive trackers (\eg, \cite{danelljan2017eco,nam2016learning}). Moreover, it is still challenging to tune the camera control module even with a strong passive tracker. Thus, we did not employ the latest passive trackers but the selected ones in this study for comparison.
We implement them by directly invoking the interface from OpenCV \cite{OpenCV} (MIL, KCF, TLD and Meanshift trackers) and Dlib \cite{Dlib} (Correlation tracker).

To make the comparison feasible, we add to the passive tracker an additional PID-like module for the camera control, enabling it as a simulated active tracker to interact with the environment (see Fig. \ref{fig:pipeline}, Right). 
Specifically, in the first frame, a manual bounding box must be given to indicate the object to be tracked. For each subsequent frame, the passive tracker then predicts a bounding box, which is passed to the ``Camera Control" module. Finally, the action is produced by ``pulling back" the target to its position in a previous frame (see supplementary materials for the details of the implementation). For a fair comparison with the proposed active tracker, we employ the same action set $\mathcal{A}$ as described in Sec. \ref{sec:approach}.

%In the first frame, a manual bounding box must be given to indicate the object to be tracked. 
%For each subsequent frame, the passive tracker then predicts a bounding box, which is passed to the ``Camera Control'' module.
%Finally, the action is produced by ``pulling back'' the target to its position in a previous frame (see supplementary materials for the details of the implementation).
%For a fair comparison with the proposed active tracker, we employ the same action set $\mathcal{A}$ as described in Sec. \ref{sec:approach}.

Armed with this camera-control module, the performance of traditional trackers is compared with the active tracker in \emph{Standard}, \emph{SharpTurn}, and \emph{Cacodemon}. 
The results in Table \ref{table:comparison_thirdparty} show that the end-to-end active tracker beats the simulated ``active'' trackers by a significant gap.
We investigate the tracking process of these trackers and find that they lose the target soon.
The Meanshift tracker works well when there is no camera shift between continuous frames, while in the active tracking scenario it loses the target soon.
Both KCF and Correlation trackers seem not capable of handling such a large camera shift, so they do not work as well as the case in passive tracking.
The MIL tracker works reasonably in the active case, while it easily drifts when the object turns suddenly. The TLD tracker works fairly well in some cases (\eg, \emph{Cacodemon}) while it is still not competitive with the proposed active tracker.

Recalling Fig. \ref{fig:sharpturn}, another reason of our tracker beating the traditional trackers is that our tracker can quickly discover the target again in the case that it is missed.
On the other hand, the simulated active trackers can hardly recover from failure cases.

\begin{figure*}[thb]
\begin{center}
\includegraphics[width=1.0\linewidth]{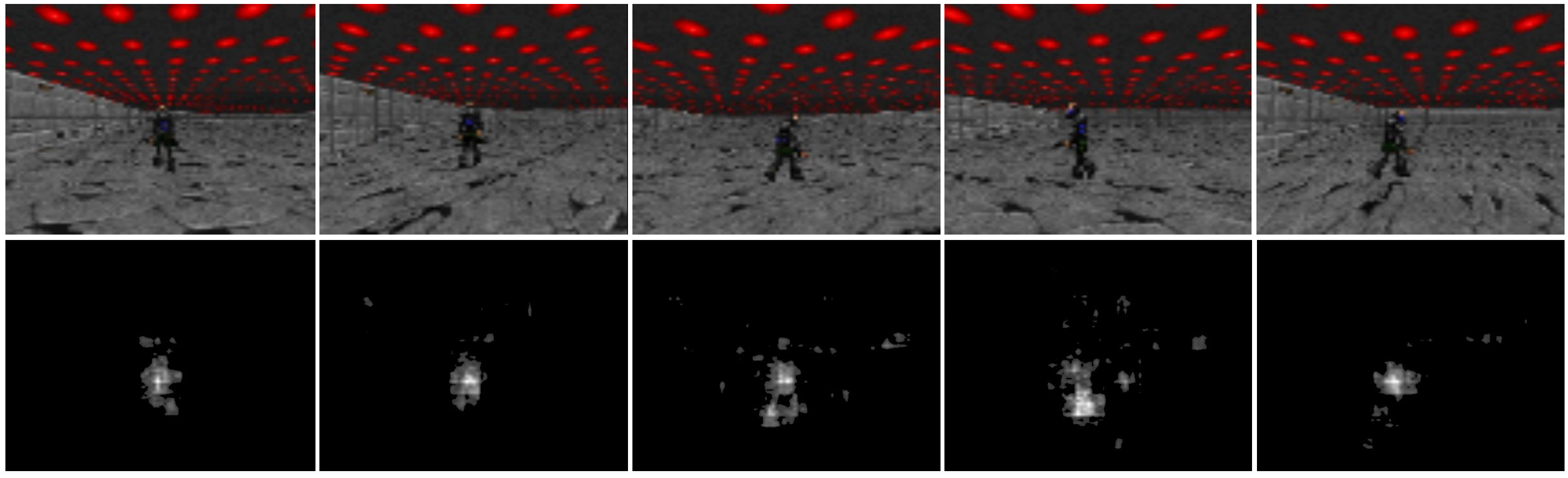}
\end{center}
\caption{Saliency maps learned by the tracker. The top row shows input observations, and the bottom row shows their corresponding saliency maps. The corresponding actions of these input images are \emph{turn-right-and-move-forward}, \emph{turn-left-and-move-forward}, \emph{turn-left-and-move-forward}, \emph{turn-left-and-move-forward}, and \emph{move-forward}, respectively. These saliency maps clearly show the focus of the tracker.}
\label{fig:saliency}
\end{figure*}

\begin{figure}[t]
\includegraphics[width=1.0\linewidth]{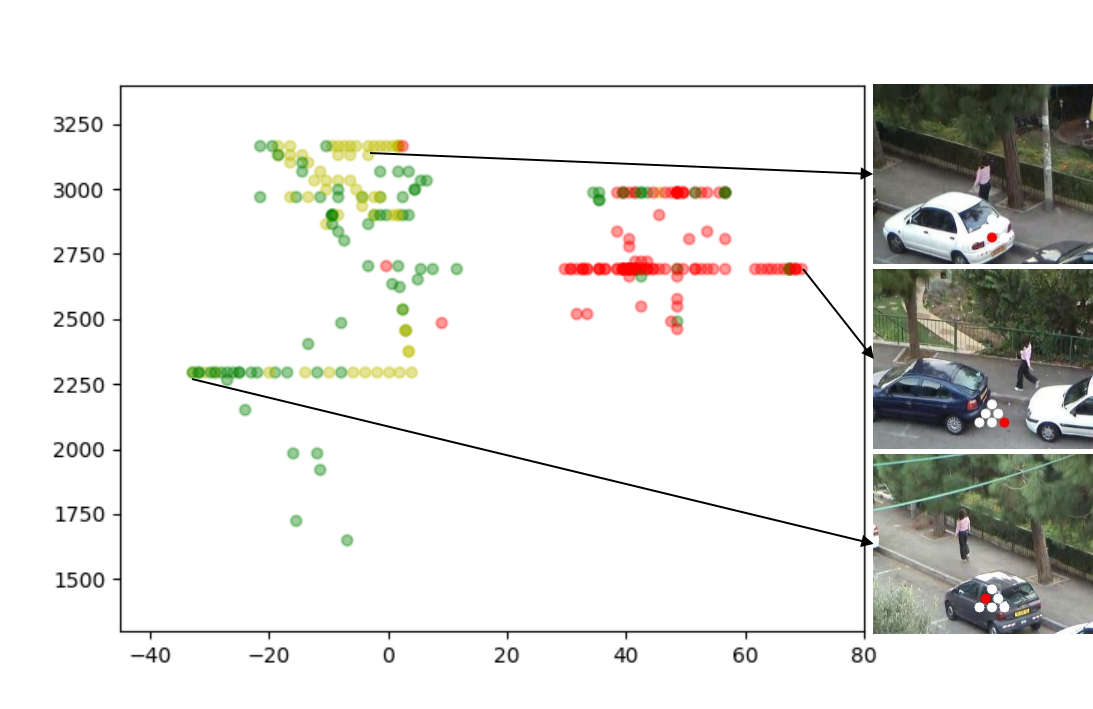}
\includegraphics[width=1.0\linewidth]{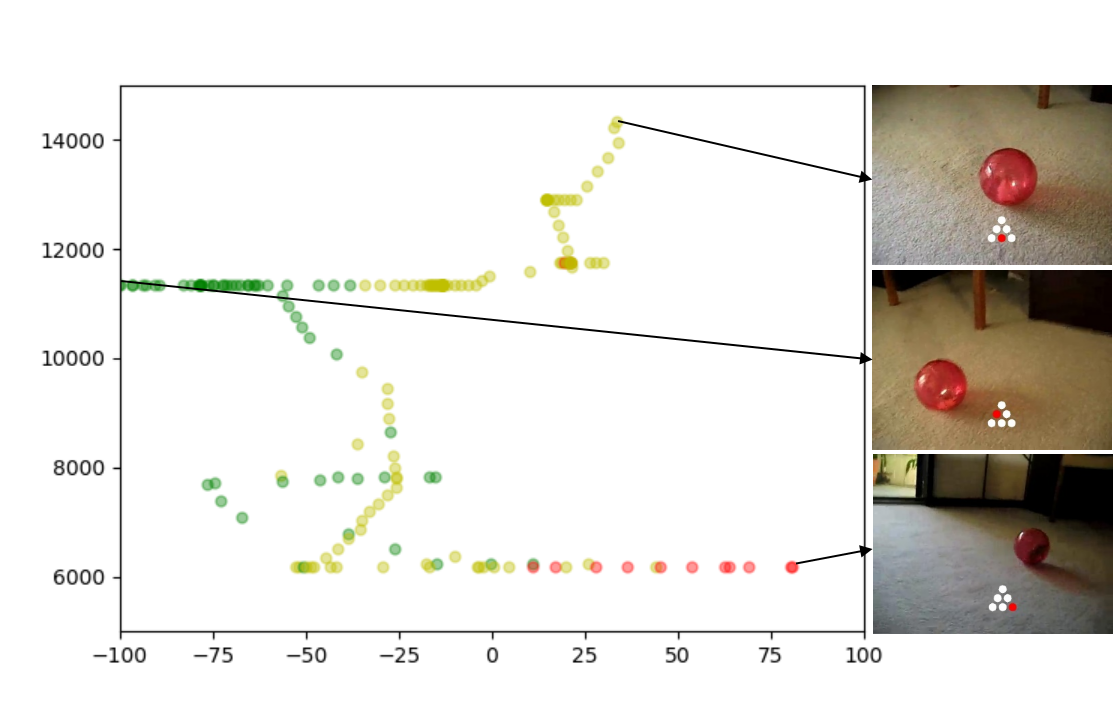} 
\caption{The output actions of the proposed active tracker in the Woman (top) and Sphere (bottom) sequences from VOT dataset.}
\label{fig:woman_sphere}
\end{figure}

The output actions of the proposed active tracker in the Woman (top) and Sphere (bottom) sequences from VOT dataset.

\begin{figure*}[t]
\hspace*{0.06\linewidth} Forward \mySkip{0.2} Left  \mySkip{0.2} Right \mySkip{0.2} Stop \\
\includegraphics[width=\picwFour]{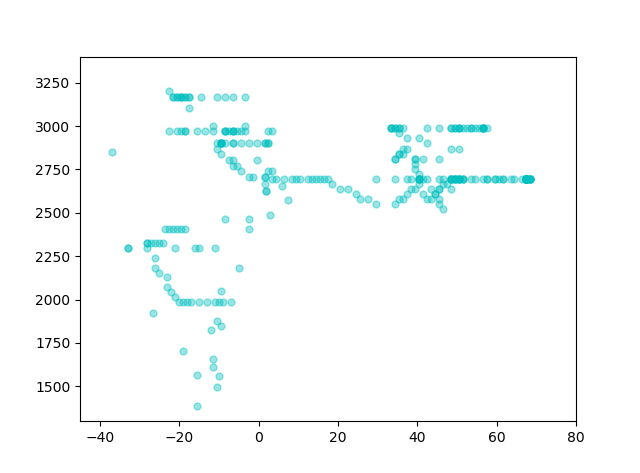}
\includegraphics[width=\picwFour]{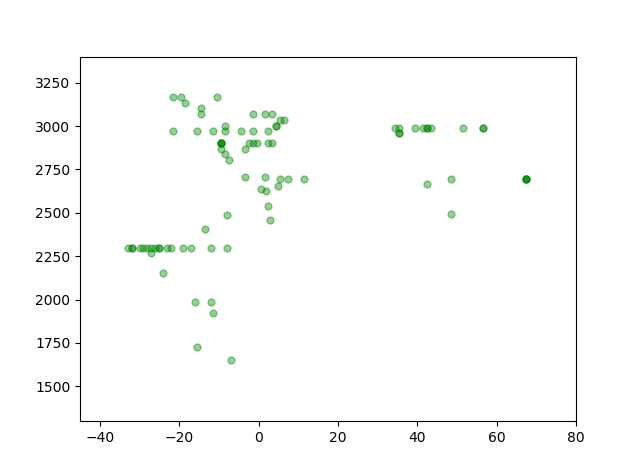}
\includegraphics[width=\picwFour]{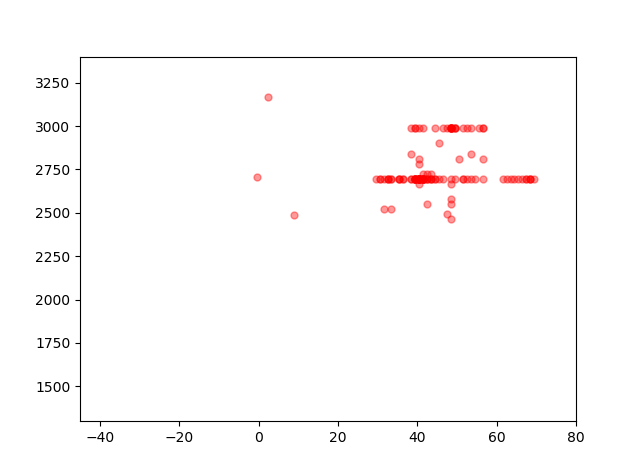}
\includegraphics[width=\picwFour]{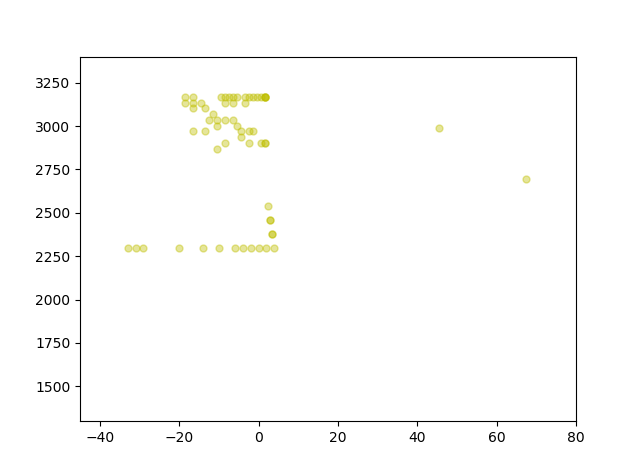} 
\caption{Visual results of individual actions of the proposed active tracker on the video clip of Woman. From left to right, they are actions of Forward (move-forward), Left (turn-left/turn-left-and-move-forward), Right (turn-right/turn-right-and-move-forward), and Stop (no-op).}
\label{fig:woman}
\end{figure*}

\begin{figure*}[th]
\hspace*{0.06\linewidth} Forward \mySkip{0.2} Left  \mySkip{0.2} Right \mySkip{0.2} Stop \\
\includegraphics[width=\picwFour]{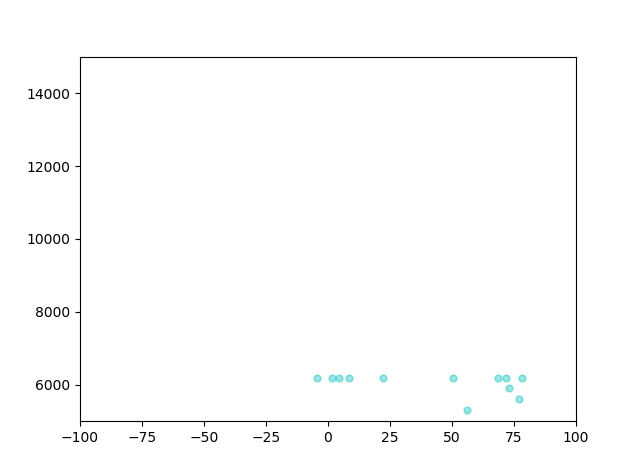}
\includegraphics[width=\picwFour]{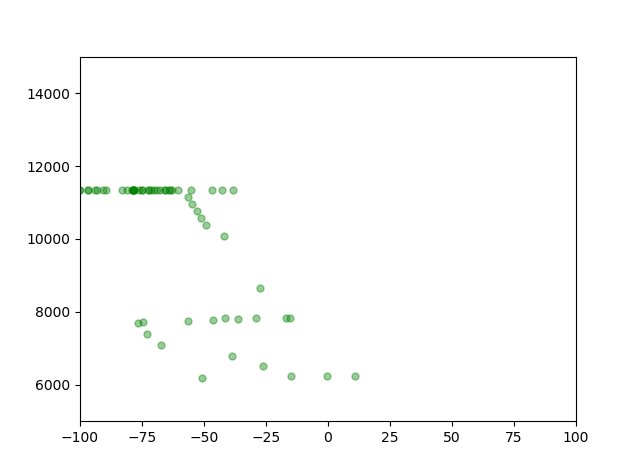}
\includegraphics[width=\picwFour]{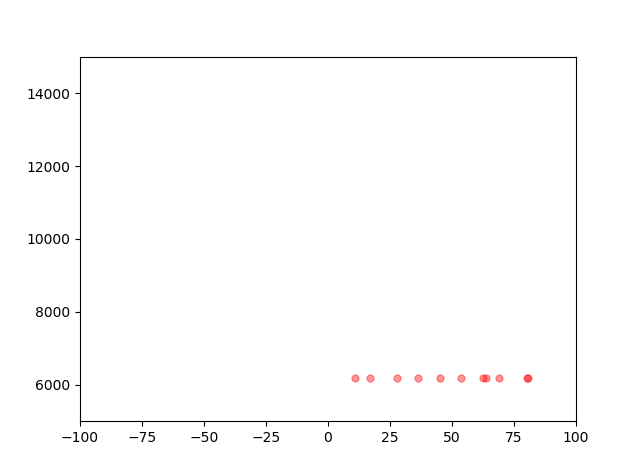}
\includegraphics[width=\picwFour]{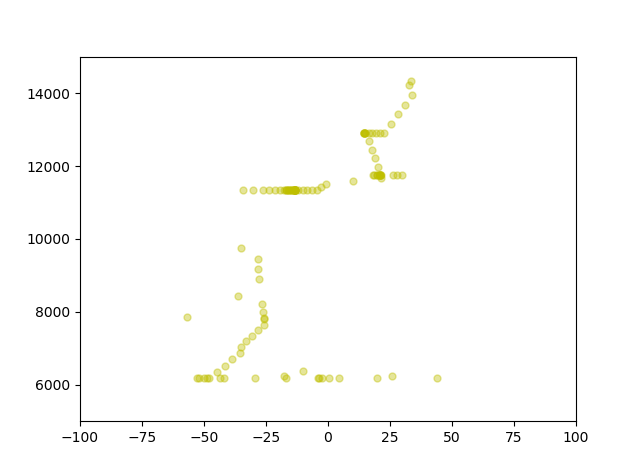} 
\caption{Visual results of individual actions of the proposed active tracker on the video clip of Sphere. From left to right, they are actions of Forward (move-forward), Left (turn-left/turn-left-and-move-forward), Right (turn-right/turn-right-and-move-forward), and Stop (no-op).}
\label{fig:sphere}
\end{figure*}

%%%%%%%%%%%%%%%%%%%%% Action Saliency Map %%%%%%%%%%%%%%%%%
\textbf{Action Saliency Map.}
\label{subsec:saliency}
We are curious about what the tracker has learned so that it leads to good performance.
To this end, we follow the method in \cite{simonyan2013deep} to generate a saliency map of the input image with regard to a specific action.
Making it more specific, an input frame $o_i$ is fed into the tracker and forwarded to output the policy function.
An action $a_i$ will be sampled subsequently. Then the gradient of $a_i$ with regard to $o_i$ is propagated backwards to the input layer, and a saliency map is generated. 
This process calculates exactly which part of the original input image influences the corresponding action with the greatest magnitude.

Note that the saliency map is image specific, \ie, for each input image a corresponding saliency map can be derived. 
Consequently, we can observe how the input images influence the tracker's actions. 
Fig. \ref{fig:saliency} shows  a few pairs of input image and corresponding saliency map. 
The saliency maps consistently show that the pixels corresponding to the object dominate the importance to actions of the tracker. 
It indicates that the tracker indeed learns how to find the target.

%%%%%%%%%%%%%%%%%%%%%% Active Tracking in the UE Environment %%%%%%%%%%%%%%%%%%%%%%
\subsection{Active Tracking in The UE Environment}
\label{subsec:ue_exp}
We first compare models trained with randomized environment and single environment.
Then we test our active tracker in four environments and also compare it against traditional trackers.

\textbf{RandomizedEnv versus SingleEnv.}
Based on the \emph{Square} environment, we train two models individually by the \emph{RandomizedEnv} protocol (random number of invisible background objects and starting point) and \emph{SingleEnv} protocol (fixed environment).
They are tested in the \emph{S2MP2} environment, where the map, target, and the path are unseen during training.
As shown in Table~\ref{table:diff_protocol_ue}, similar results are obtained as those in Table~\ref{table:diff_protocol}.
We believe that the improvement benefits from the environment randomness brought by the proposed environment augmentation techniques.
In the following, we only report the results of the \emph{RandomizedEnv} protocol.

\begin{table}
\caption{Performance under different protocols in \emph{S2MP2}.}
\label{table:diff_protocol_ue}
\centering
{\begin{tabular}{c c c c}
\hline
\textbf{Protocol} & \textbf{AR} & \textbf{EL}\\
\hline
\emph{RandomizedEnv}   &1203.6$\pm$1428.4 & 2603.6$\pm$503.0 \\
\emph{SingleEnv}    & -453.4$\pm$1.5  & 461.9$\pm$180.0 \\
\hline
\end{tabular}
}
\end{table}

\textbf{Various Testing Environments.} 
To intensively investigate the generalization ability of the active tracker, we test it in four different environments and present the results in Table \ref{table:comparison_thirdparty_ue}. 
We compare it with the simulated active trackers described in Sec. \ref{subsec:vizdoom_exp}, as well as one based on the long-term TLD tracker \cite{kalal2012tracking}.

According to the results in Table \ref{table:comparison_thirdparty_ue} we conduct the following analysis:
1) Comparison between \emph{S1SP1} and \emph{S1MP1} shows that the tracker generalizes well even when the model is trained with target Stefani, revealing that it does not overfit to a specialized appearance.
2) The active tracker performs well when changing the path (\emph{S1SP1} versus \emph{S1SP2}), demonstrating that it does not act by memorizing a specialized path.
3) When we change the map, target, and path at the same time (\emph{S2MP2}), though the tracker could not seize the target as accurately as in previous environments (the AR value drops), it can still track objects robustly (comparable EL value as in previous environments), proving its superior generalization potential.
4) In most cases, the proposed tracker outperforms the simulated active tracker, or achieves comparable results if it is not the best.
The results of the simulated active tracker also suggest that it is difficult to tune a unified camera-control module for them, even when a long term tracker is adopted (see the results of TLD). 
However, our work exactly sidesteps this issue by training an end-to-end active tracker.

\begin{table}[t]
\caption{Comparison with traditional trackers. The best results are shown in bold.}
\label{table:comparison_thirdparty_ue}
\centering
{\begin{tabular}{c  c  c  c}
\hline
\textbf{Environment} & \textbf{Tracker} & \textbf{AR} & \textbf{EL}\\
\hline
\multirow{5}{*}{\emph{S1SP1}}
& MIL        & -453.8 $\pm$0.8 &  601.4 $\pm$ 300.9  \\
& Meanshift  & -454.1$\pm$1.3 &  628.6$\pm$111.2    \\
& KCF         & -453.6$\pm$2.5 &  782.4$\pm$136.1  \\
& Correlation & -454.9$\pm$0.9 & 1710.4$\pm$417.0  \\
& TLD & -453.6$\pm$1.3 & 376.0$\pm$70.9  \\
\cline{2-4}
& Active   & \textbf{2495.7}$\pm$\textbf{12.4} &  \textbf{3000.0}$\pm$\textbf{0.0} \\
\hline
\multirow{5}{*}{\emph{S1MP1}}
& MIL        & -358.7$\pm$189.4 &  1430.0$\pm$825.3   \\
& Meanshift   & 708.3$\pm$3.3 &  \textbf{3000.0}$\pm$\textbf{0.0}   \\
& KCF         & -453.6$\pm$2.6 & 797.4$\pm$37.5    \\
& Correlation & -453.5$\pm$1.3 & 1404.4$\pm$131.1  \\
& TLD & -453.4$\pm$2.0 & 651.0$\pm$54.5  \\
\cline{2-4}
& Active   & \textbf{2106.0}$\pm$\textbf{29.3}  & \textbf{3000.0}$\pm$\textbf{0.0} \\
\hline
\multirow{5}{*}{\emph{S1SP2}}
& MIL         & -452.4$\pm$0.7 &  420.2$\pm$104.9   \\
& Meanshift   & -453.0$\pm$1.8  &    630.2$\pm$223.8 \\
& KCF         & -453.9$\pm$1.5 &  594.0$\pm$378.8   \\
& Correlation & -452.4$\pm$0.4 &  293.8$\pm$97.4  \\
& TLD & -454.7$\pm$1.8 & 218.0$\pm$26.0  \\
\cline{2-4}
& Active   & \textbf{2162.5}$\pm$\textbf{48.5}  & \textbf{3000.0}$\pm$\textbf{0.0}  \\

\hline
\multirow{5}{*}{\emph{S2MP2}}
& MIL         & -453.1$\pm$0.9 &  749.0$\pm$301.0   \\
& Meanshift   & 726.5$\pm$10.8  &  \textbf{3000.0}$\pm$\textbf{0.0}   \\
& KCF         & -452.4$\pm$1.0 &  247.8$\pm$18.8   \\
& Correlation & -215.0$\pm$475.3 & 1571.6$\pm$919.1  \\
& TLD & -453.1$\pm$1.8 & 208.8$\pm$33.1  \\
\cline{2-4}
& Active   &  \textbf{740.0}$\pm$\textbf{577.4} & 2565.3$\pm$339.3  \\
\hline
\end{tabular}
}
\end{table}

\subsection{Transfer Potential in The VOT Dataset}
\label{subsec:vot_exp}
To evaluate how the active tracker would perform in real-world scenarios, we take the network trained in a UE4 environment with an environment augmentation method and test it on a few video clips from the VOT dataset \cite{VOT_TPAMI}. 
Obviously, we can by no means control the camera action for a recorded video. 
However, we can feed in the video frame sequentially and observe the output action predicted by the network, checking whether it is consistent with the actual situation.

Fig. \ref{fig:woman_sphere} shows the output actions for two video clips named Woman and Sphere, respectively. 
The horizontal axis indicates the position of the target in the image, with a positive (negative) value meaning that a target in the right (left) part. 
The vertical axis indicates the size of the target, \ie, the area of the ground truth bounding box. Green and red dots indicate turn-left/turn-left-and-move-forward and turn-right/turn-right-and-move-forward actions, respectively. 
Yellow dots represent No-op action. Figures \ref{fig:woman} and \ref{fig:sphere} show actions individually according to our discrete actions. The actions are grouped as Forward (Move-forward), Left (including both Turn-left and Turn-left-and-move-forward actions in our action space), Right (including both Turn-right and Turn-right-and-move-forward actions in our action space), and Stop (No-op).
As the figures show, 1) When the target resides in the right (left) side, the tracker tends to turn right (left), trying to move the camera to ``pull'' the target to the center. 
2) When the target size becomes bigger, which probably indicates that the tracker is too close to the target, the tracker outputs no-op actions more often, intending to stop and wait the target to move farther. 

We believe that the qualitative evaluation shows evidence that the active tracker, learned from a purely virtual environment, could reasonably map the real-world observation to appropriate actions. However, as mentioned before, the video is ``passive'' and could not be controlled by the tracker. The setting of this experiment does not strictly follow that of active tracking. Thus the practical value still has uncertainty.
In Sec.~\ref{subsec:real-world-exp}, we conduct experiments to deploy the active tracker in a real-world robot to further demonstrate the generalization of the tracker learned in the virtual environment.

%%%%%%%%%%%%%%%%%%%%% Active Tracking in Real-world Scenarios %%%%%%%%%%%%%%%%%
\subsection{Active Tracking in The Real-world Scenarios}
\label{subsec:real-world-exp}
We deploy our tracker in a real-world robot and test it in both indoor and outdoor scenarios to verify the practical value of the proposed active tracker. 

\subsubsection{Training}
\label{subsec:real-world-exp-train}
In general,
it is not straightforward to deploy an agent trained purely in virtual environment to the real world due to the existence of the virtual-to-real gap. 
In order for a successful deployment, 
we have trained the active tracker with more advanced environment augmentation techniques and more appropriate action space, 
which will be elaborated as follows.   

\begin{figure*}[th]
\centering
\includegraphics[width=0.95\linewidth]{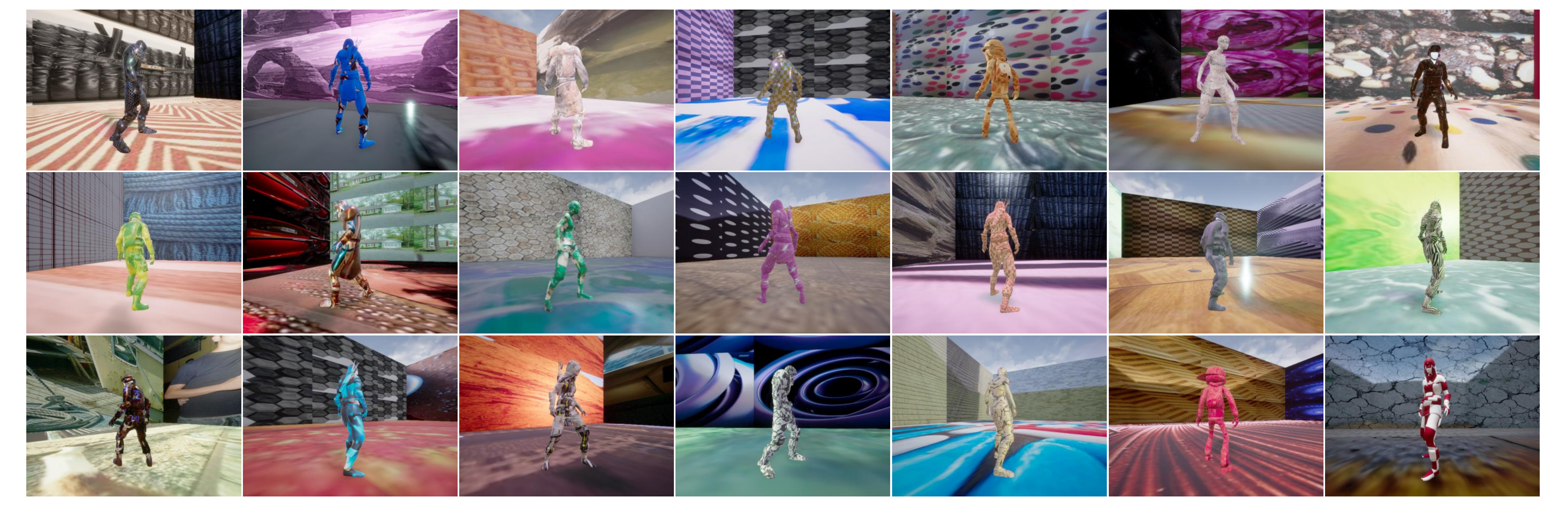}
\caption{Examples of rendered images using the texture randomization in the augmented environment for training the tracker.}
\label{fig:random_env}
\end{figure*}

% Env augmentation
\textbf{More Advanced Environment Augmentation.}
%We believe the tracker trained with UE environment should be adequate for deployment in the real world, while the diversity of training environment should be boosted at first to benefit the generalization of the tracker. 
By taking advantage of the UE environment augmentation,
we believe that training the tracker in UE simulator alone is sufficient for a successful real-world deployment.
No fine-tuning in the real world is needed.
Towards this goal,
we further extend the environment augmentation technique in Sec. \ref{sec:aug-env} by randomizing more aspects of the environment during training, 
including the texture of each mesh, the illumination condition of the scene, the trajectory of the target, as well as the speed of the target. 
Specifically,
1) for the textures, 
we randomly choose pictures from two image datasets~\cite{Kylberg2011c, zhou2017places} and place them on the surface of each background object and the target object, shown as Fig.~\ref{fig:random_env}.
2) For the illumination condition, 
we randomize the intensity and color of each light source in the environment as well as each position, orientation.
3) For the moving trajectory and the velocity of the target, 
we randomly sample a coordinate in the training environment as the goal of the target, and generate a trajectory using the built-in navigation module of the UE engine, 
which can ensure that the target moves to the goal and avoids any obstacle.
The speed of the target during its movement to the current goal is randomly sampled in the range of $(0.1 m/s, 1.5 m/s)$. 

With the texture and illumination randomizations, the trained tracker can avoid over-fitting to specific appearance of targets and backgrounds, 
and learn to infer ``\emph{what}" target to be tracked. 
The trajectory and speed randomization can guide the sequence encoder to learn the inference of ``\emph{how}" the target is moving and to implicitly encode the motion-related feature, 
which makes it easier to adapt the actor network to various motion patterns.

\textbf{More Appropriate Action Space.}
The original action space includes six discrete actions, 
which are inadequate for the deployment of a real robot 
in the sense that the tracker cannot adapt to different moving speeds of the target.
% A typical observation when we deploy the active tracker with the original action space is that, the tracker cannot adapt to different moving speed of the target. 
To this end, we extend the original six discrete actions to nine discrete actions.
The enhanced discrete action space additionally includes the backward action, 
which enables the robot to move backwards when the target is too close to the agent.
Moreover, this improved discrete action space also enables the agent to move backward and forward with two different levels of speed. 
%Each discrete action corresponds to a configuration of the linear and angular velocities of the agent (the second column in Table~\ref{table:action_mapping_discrete}) and is mapped to the linear and angular velocities of the real robot (the third column in Table~\ref{table:action_mapping_discrete}).
In Table~\ref{table:action_mapping_discrete} we give the nine actions,
each with the configured linear and angular velocity in both the simulator (the second column) and real-world robot (the third column).

%We also curious about the performance with continuous action space, thus we also investigate the active tracker with continuous action space. 
We are also curious about the performance of adopting a continuous action space.
We thus investigate a two-dimensional action space composing linear velocity and angular velocity.
Akin to the configuration of the discrete actions, we also well configure the linear and angular velocities in both the simulator and the robot, 
as in Table~\ref{table:action_mapping_continuous},
where we also show the value range for each dimension.

\begin{table}[t]
\caption{Mapping discrete actions from virtual to real. The second and the third columns are the linear and angular velocities in the virtual and the real robot, respectively.}
\label{table:action_mapping_discrete}
\centering
{\begin{tabular}{c  c  c }
\hline
\multirow{2}{*}{\textbf{Discrete Action}}
& Linear (cm/step),  & Linear (m/s),  \\
& Angular (degree/s) & Angular (rad/s) \\
\hline
Forward (fast) & 50, 0 & 0.4, 0 \\
Forward (slow) & 25, 0 & 0.2, 0  \\
Backward (fast) & -50, 0 & -0.4, 0 \\
Backward (slow) & -25, 0 & -0.2, 0 \\
Turn Left & 0, 10     & 0, 0.6  \\
Turn Right & 0, -10    & 0, -0.6 \\
Turn Left \& Forward & 15, 5  & 0.1, 0.2  \\
Turn Right \& Forward & 15, -5 & 0.1, -0.2 \\
Stop      & 0, 0     & 0,  0 \\
\hline
\end{tabular}
}
\end{table}

\begin{table}[t]
\caption{Mapping continuous actions from virtual to real. The second and the third columns are the value ranges of velocities in the virtual and the real robot, respectively.}
\label{table:action_mapping_continuous}
\centering
{\begin{tabular}{c  c  c }
\hline
\multirow{2}{*}{\textbf{Bound of Action}}
& Linear (cm/step),  & Linear (m/s),  \\
& Angular (degree/s) & Angular (rad/s) \\
\hline
High & 80, 20 & 0.4, 0.6   \\
Low & -80, -20 & -0.4, -0.6  \\

\hline
\end{tabular}
}
\end{table}

\subsubsection{Robot Setting}
We employ a TurtleBot to perform the deployment experiment. 
As shown in Fig. \ref{fig:robot}(a),
it is a wheeled robot 
equipped with an RGB-D camera that is amounted at the height about 80cm. 
The robot is controlled to move via sending commands of expected linear velocity and angular velocity. 
We use a laptop with 8-core Intel CPU as our platform to conduct the task, including acquiring images from the camera, predicting action via the neural network, and sending the action command to the controller. The state and action of the tracker are updated every 50ms (20Hz).

% robot experiment setup
\subsubsection{The Real-World Testing Environment}
To investigate how successful the deployment is, we test the deployed active tracker with the robot configured above in two
scenarios: \emph{indoor room} and \emph{outdoor rooftop}. In each testing scenario, the pedestrian is taken as the object to be tracked.

\begin{figure*}[th]
\hspace*{0.01\linewidth}(a) \mySkip{0.23} (b) \mySkip{0.37} (c) \\
\centering
\includegraphics[width=0.15\linewidth]{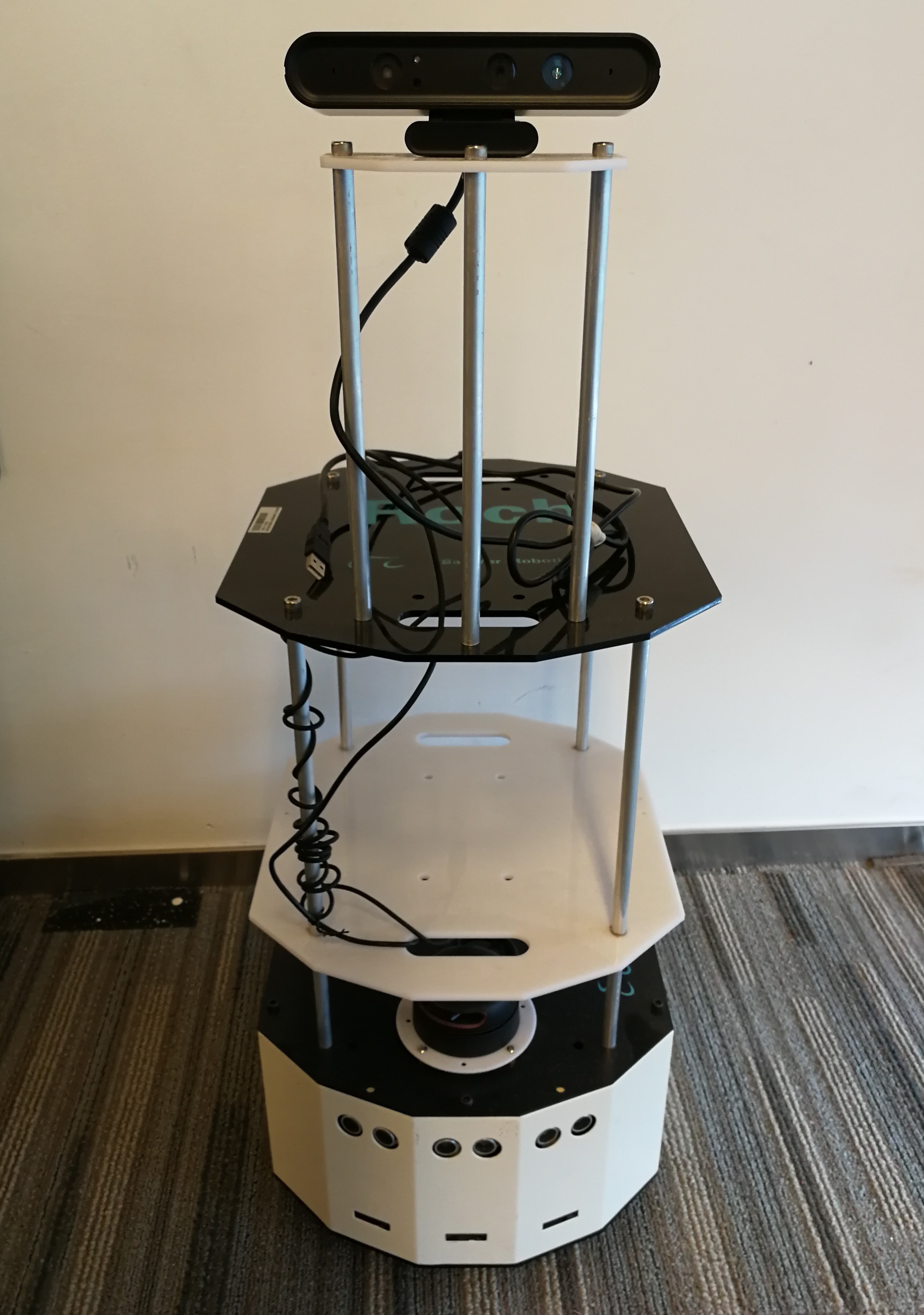}
\includegraphics[width=0.38\linewidth]{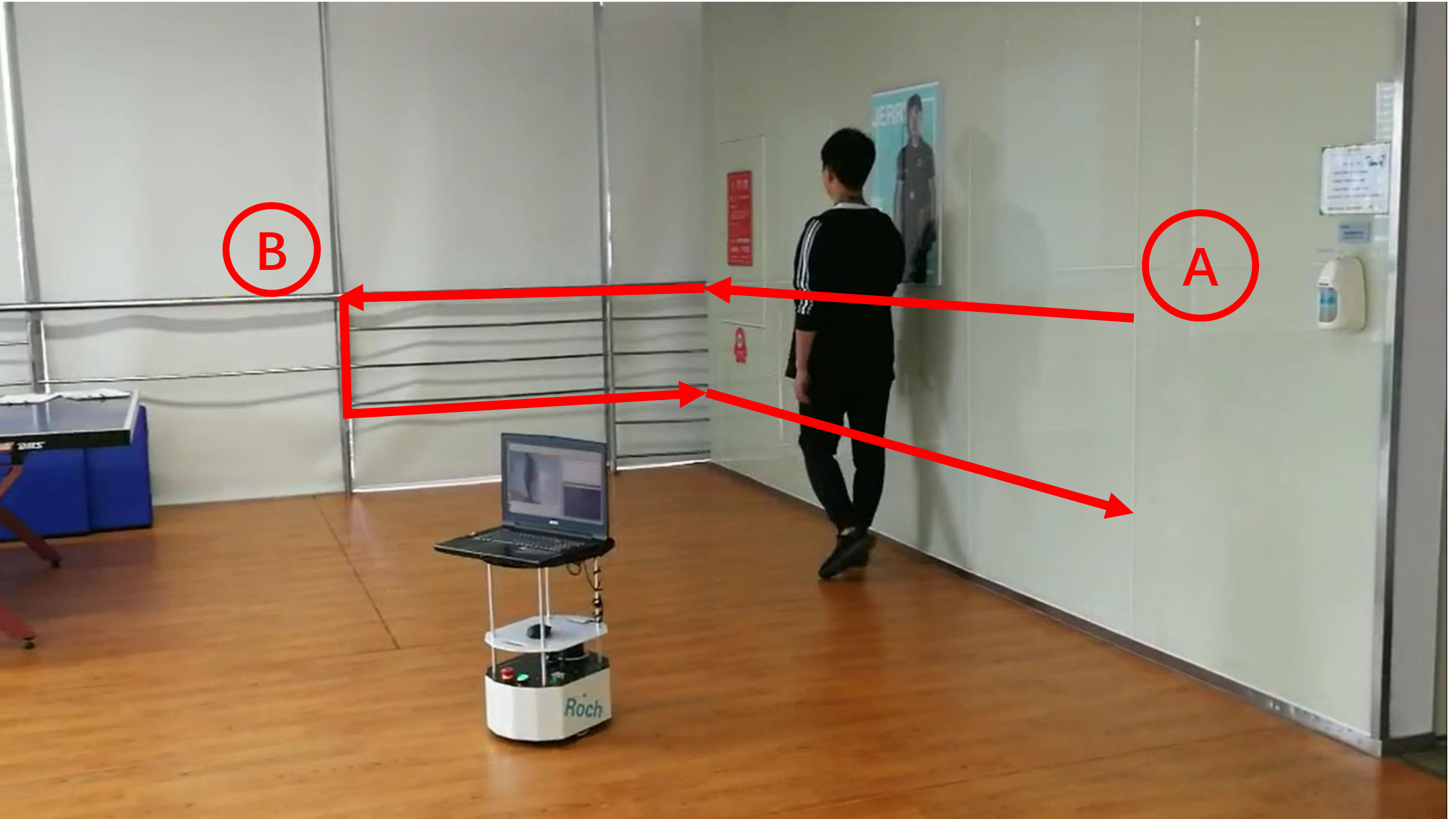}
\includegraphics[width=0.38\linewidth]{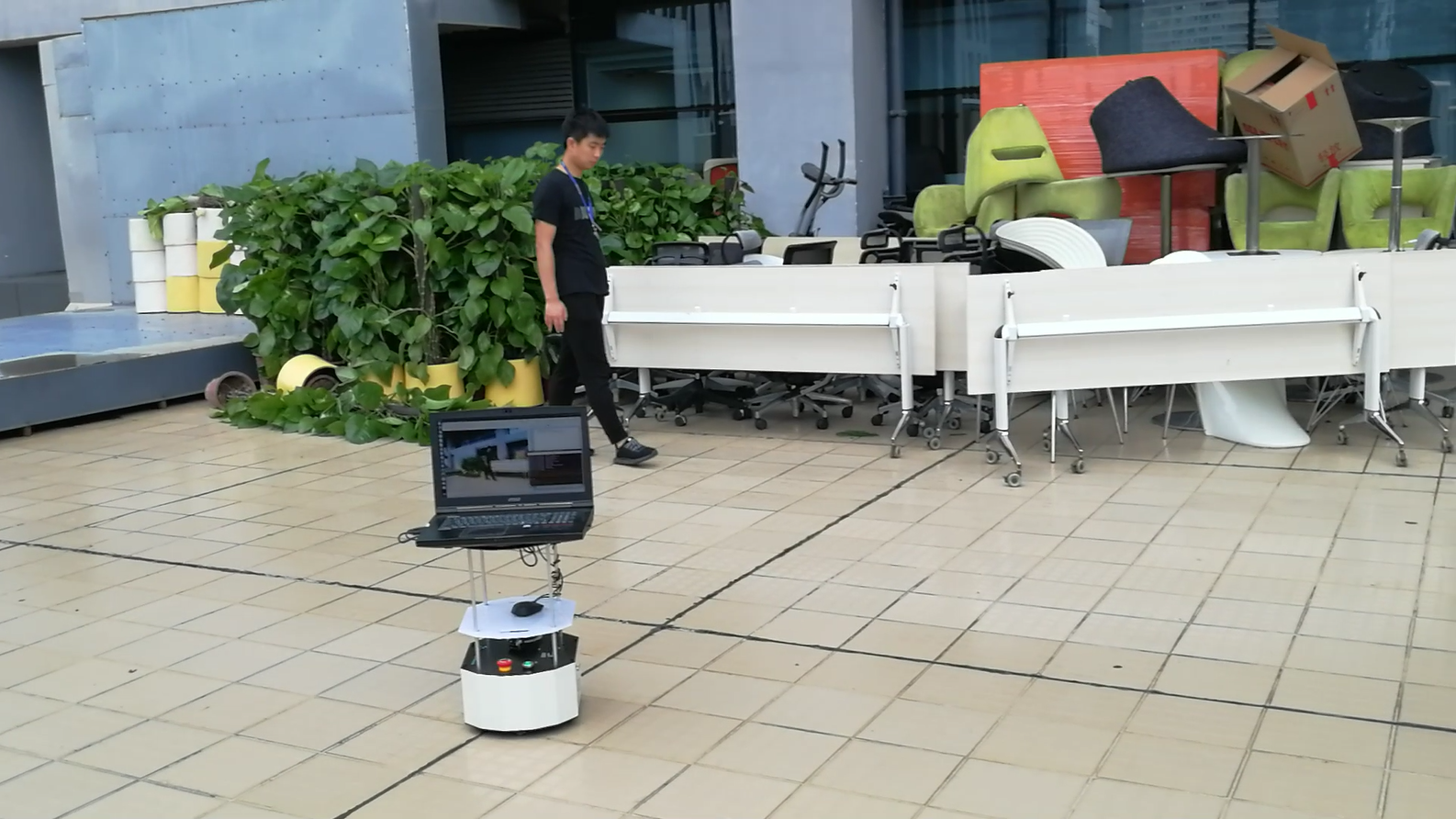}
\caption{\textbf{The setup of the deployment experiments.} a) The robot used in our experiment. b) A snapshot of the \emph{indoor room} environment used for evaluation. c) A snapshot of the \emph{outdoor rooftop} environment used for evaluation. }
\label{fig:robot}
\end{figure*}

In \emph{indoor room}, there are a table, a reflective glass wall, and a row of railings. 
A snapshot is shown in Fig. \ref{fig:robot}(b).
The reflective glass wall makes the texture dynamically change while viewed from different positions, which may distract the tracker.
Besides, there is a man in the poster on the wall, who is similar to the object. This may further distract the tracker.

In \emph{outdoor rooftop}, there are buildings, desks, chairs, plants, and sky, as shown in a snapshot of Fig. \ref{fig:robot}(c).
Compared to the \emph{indoor room}, the background in \emph{outdoor rooftop} is much more complicated (see the randomly piled up a set of objects, including desks, chairs, and plants).
And in this scenario it is difficult for the camera to expose accurately, due to the uneven illumination conditions.
Additional challenges are posed for the tracker to perceive the target in the inaccurately exposed observation image, such as frames \#$0$ and \#$451$ in Fig.~\ref{fig:outdoor_seq}.
%The directional light caused by sun makes it challenge. 
%It make the appearance of target look different from different orientation.
%When tracker looks at the target in the backlight, it becomes a black shadow.
%It also makes the background observed in high dynamic ranges.

%It is more convincing to test the active tracker in multiple episodes. 
To make the results more convincing, we test the active tracker in multiple episodes.
%To make the testing environment as consistent as possible in each episode, 
For consistency,
the object is required to follow a specific trajectory in each testing environment.
For example, the person walks along the wall from point A to point B and then goes back to A in the environment of \emph{indoor room}, shown as the red line in Fig.~\ref{fig:robot}(b). 
And the tracker also starts from a specific starting point.

\subsubsection{Quantitative Results}
\label{subsec:real-world-exp-quant}
% \textbf{Metrics}
To evaluate the performance quantitatively, we consider the perspective of both \emph{robustness} and \emph{accuracy}.
For robustness, we count the number of successful and failure episodes in both testing scenarios.
Unlike in simulator,
it is impossible to obtain the reward signal in the real-world environment,
as it is too expensive to obtain the coordinate of the tracker/target.
We then consider that the tracker follows the target successfully only when the target appears apparently in the observed image. 
During the course, once the target is lost for continuous three seconds, the episode is done, and we count such an episode as a \emph{failure}.
If the tracker keeps tracking until the target finishes a specific trajectory, the episode is also done, and we count such an episode as a \emph{success}. 
For accuracy, we consider the consistency of the target size and the deviation between the target and the center of the image.  
Specifically, we apply a state-of-the-art object detector, YOLOv3~\cite{redmon2018yolov3}, to detect the target and obtain its bounding box.
Based on the bounding box, 
we calculate the variation of the size of the target and the deviation between the target center and the image center during tracking.
The two metrics indicate how accurately the tracker follows the target.
Note that, to rule out the effect of different image sizes, 
we calculate the two metrics in a relative way, \ie, 
the target size is defined as box size over image size,
and the deviation is defined as absolute deviation (could be negative) over half image width.

We test the tracker with both the enhanced discrete actions and continuous actions. 
Table~\ref{table:comparison_real} summarizes the performance of the tracker in both testing environments.
$10$ episodes are performed in each scenario.
In \emph{indoor room}, both the discrete and continuous trackers successfully follow the target during most of the testing episodes.
Comparing the numbers of these two trackers, 
the discrete action tracker successfully accomplishes the total ten episodes without failures, 
but the continuous tracker fails three times out of ten episodes.
The variance of the target size does not show any difference, while the
deviation of the action for the discrete tracker is smaller than that of the continuous tracker.
A similar observation can be obtained when we analyze the results in the \emph{outdoor rooftop} environment.
This suggests that the tracker is more robust with the enhanced discrete action space.
We conjecture that the space discretization makes the agent more robust to the noises from both the background distraction and the robot control system.

\begin{table}[t]
\caption{Comparison between active trackers with different types of action space. The target size and the deviation are relative values considering the full image size.}
\label{table:comparison_real}
\centering
{\begin{tabular}{c  c  c  c c}
\hline
\textbf{Env} & \textbf{Action Space} & \textbf{Success Rate} & \textbf{Target Size} & \textbf{Deviation}\\
\hline
\multirow{2}{*}{\emph{Indoor}}
& Discrete        & 1.0 &  0.14$\pm$0.09 &  0.03$\pm$0.20  \\
& Continuous   & 0.7 &  0.12$\pm$0.09   &  0.08$\pm$0.09 \\
\hline
\multirow{2}{*}{\emph{Outdoor}}
& Discrete        & 0.9 &  0.15 $\pm$0.06 &  -0.02$\pm$0.26  \\
& Continuous   & 0.7 &  0.17 $\pm$0.09   &  -0.09$\pm$0.29 \\
\hline
\end{tabular}
}
\end{table}

\subsubsection{Qualitative Results}
\label{real-world-exp-qual}
\begin{figure*}[th]
\centering
\includegraphics[width=0.95\linewidth]{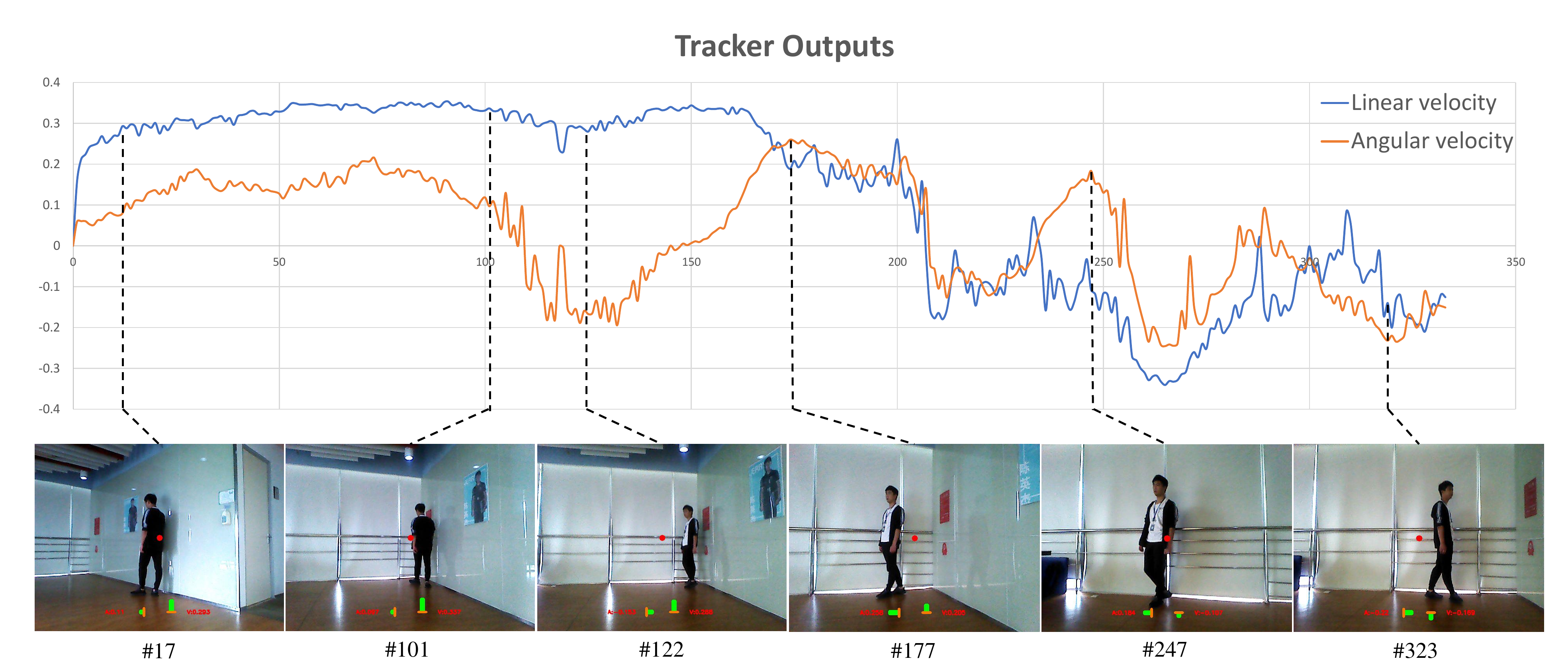}
\caption{A demo sequence about our continuous active tracker performing in \emph{indoor room}.
Note that the value polarity corresponds to different directions of velocity.
As for angular velocity, the positive is left, and the negative is right .
As for linear velocity, the positive corresponds to forward, and the negative corresponds to backward.
The red dot in the image center is a reference point to help us mark the relative location between the object and robot.
The length of the green bar in the bottom represents the magnitude of velocity.
The horizontal bar represents the angular velocity, and the vertical bar represents the linear velocity.
The orange line represents the zero value.
}
\label{fig:indoor_seq}
\end{figure*}

\begin{figure*}[th]
\centering
\includegraphics[width=0.95\linewidth]{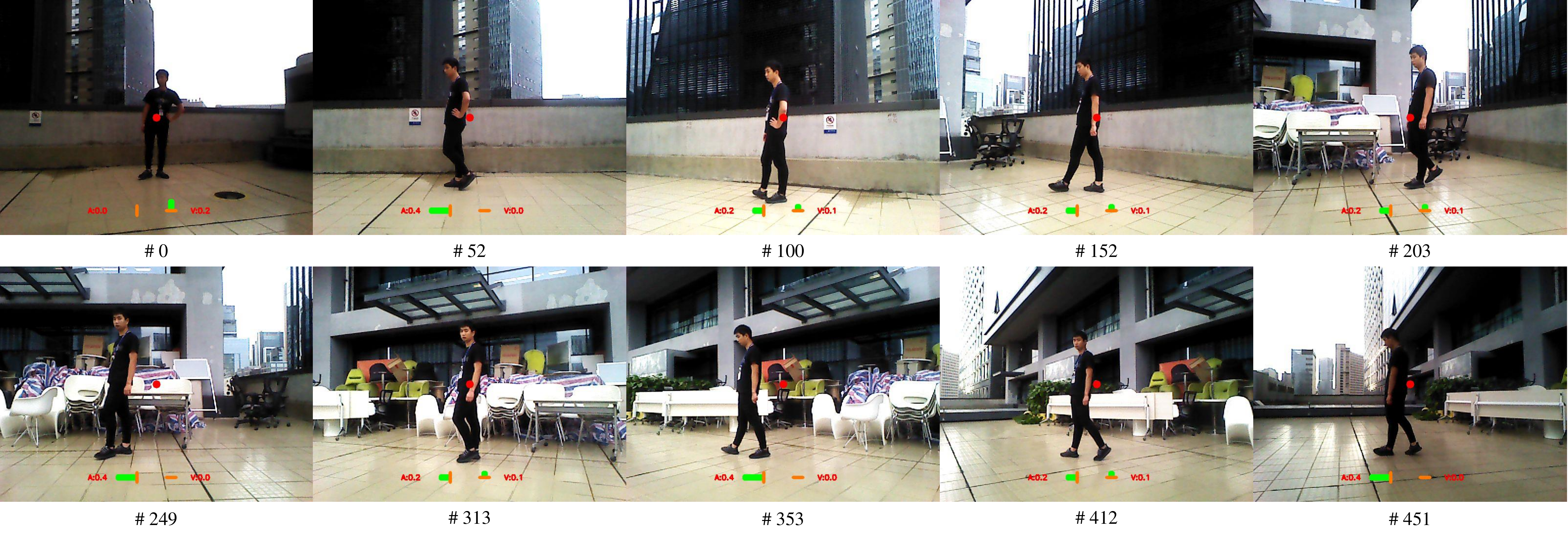}
\caption{A demo sequence about our active tracker executing discrete actions in the \emph{outdoor rooftop} environment.
The red dot is the center of the image. The horizontal bar represents the value of the expected angle velocity, and the vertical one represents the value of the expected linear velocity. The orange line is zero.
}
\label{fig:outdoor_seq}
\end{figure*}

We visualize two typical sequences from the recorded videos, 
for the continuous action tracker running in the \emph{indoor room} (Fig.~\ref{fig:indoor_seq}) environment 
and the discrete action tracker running in the \emph{outdoor rooftop} (Fig.~\ref{fig:outdoor_seq}) environment.
Both sequences illustrate that the active tracker trained in pure virtual environments from scratch is able to transfer to real-world scenarios.
No matter whether the tracker relies on discrete or continuous action spaces, 
it tends to ``place'' the target in the image center, 
and to ``keep'' the size of the target constant.

We provide an interpretation of the curves in Fig.~\ref{fig:indoor_seq}.
In the \emph{indoor room} sequence, the object starts from point A (frame \#17), walks along the wall to the corner (frame \#101), turns left (frame \#122), moves towards point B (frame \#247), and finally turns around and walks along the wall to the start point A.
At the beginning of this sequence (from frame \#1 to \#101), the tracker performs turning left and moving forward simultaneously, 
because from the perspective of the tracker the object is observed to move forward and to the left side relative to the robot coordinate system. 
After a while, the object reaches the corner and stops moving for a moment before turning to left. 
However, due to the inertia of the physical system, it is difficult for the robot to stop moving immediately while running at a high speed, even though the tracker has already realized that the target is on the right and outputs a negative angular velocity (indicating turning right), shown as frame \#122. 
When the object walks towards point B, the angular velocity begins to increase from negative (turn right) to positive (turn left) and holds in a level until the object is placed in the right of the image.
At the same time, the linear velocity is decreasing when the object size increases. 
At frame \#247, the object arrives at point B, turns around, and starts moving back to point A.
Since then, the robot begins to move backward and turn right, roughly performing the opposite of the previous actions (frame \#323).

Besides, when the moving object is close to the center of image, 
the tracker does not perform a naive ``stop'' action.
Instead,
the tracker usually keeps moving towards a specific direction, which is consistent with the moving direction of the object,
as shown in frames \#$17$, \#$101$ and \#$247$. 
We conjecture that the sequence encoder extracting the motion-related feature helps the actor network make a more reasonable action decision to control the robot.

%%%%%%%%%%%%%%%%%%%%%%%%%%%%%%%%%%%%%%%%%%%%%%%%%%%%%%%%%%%%%%%%%%%%%%%%%%%%%%%%%%%%%%
%%%%%%%%%%%%%%%%%%%%%%%%%%%%%%%%%% CONCLUSION %%%%%%%%%%%%%%%%%%%%%%%%%%%%%%%%%%%%%%%%
%%%%%%%%%%%%%%%%%%%%%%%%%%%%%%%%%%%%%%%%%%%%%%%%%%%%%%%%%%%%%%%%%%%%%%%%%%%%%%%%%%%%%%

\section{Conclusions}
In this paper, we proposed an end-to-end active tracker via deep reinforcement learning. 
Unlike conventional passive trackers, the proposed tracker is trained in simulators, saving the efforts of human labeling or trail-and-errors in the real world.
It shows good generalization to unseen environments. The tracking ability can potentially be transferred to the real-world scenarios. By developing more advanced environment augmentation techniques and using more appropriate action spaces, 
we have successfully deployed a robot that performs active tracking in the real world.

%\appendices

%\section{}

\ifCLASSOPTIONcompsoc
  \section*{Acknowledgments}
\else
  \section*{Acknowledgment}
\fi

%The authors would like to thank Gangming Zhao for taking the time to test our robot.
The authors would like to thank Jia Xu for his helpful discussion, Tingyun Yan for his help in building virtual environment, and Chao Zhang for setting up the robot in our early work.
Fangwei Zhong and Yizhou Wang were supported in part by the following grants: NSFC-61625201, NSFC-61527804, Tencent AI Lab Rhino-Bird Focused Research Program No.JR201851, Qualcomm University Collaborative Research Program.

\ifCLASSOPTIONcaptionsoff
  \newpage
\fi



\bibliographystyle{IEEEtran}
\bibliography{refs.bib}
\end{document}